\documentclass{article}

\usepackage{microtype}
\usepackage{graphicx}
\usepackage{subfigure}
\usepackage{booktabs}
\usepackage{xifthen}
\usepackage{chngcntr}
\usepackage{amsmath}
\usepackage{amssymb}

\usepackage{hyperref}

\DeclareGraphicsExtensions{.pdf}

\usepackage[accepted]{icml2018}

\def \papertitle{Graph Networks as Learnable Physics Engines for Inference and Control}

\icmltitlerunning{\papertitle}

\def \environmentrandomtrajectories{https://drive.google.com/file/d/14eYTWoH15T53a7qejvCkDLItOOE9Ve7S/view?usp=sharing}

\newcounter{videocounter}

\newcommand\video[3][]%
{%
  \stepcounter{videocounter}%
  \ifthenelse{\isempty{#1}}%
  {\expandafter\edef\csname#2name\endcsname{vid-\thevideocounter}}%
  {\expandafter\edef\csname#2name\endcsname{link-#1}}%
  \expandafter\edef\csname#2\endcsname{#3}%
}

\video[P.F.Pe]{rolloutspendulum}{https://drive.google.com/file/d/1NuO9_XnpNtwU4cFHXXvlJsv-FXGShbtH/view?usp=sharing}

\video[P.F.Ca]{rolloutscartpole}{https://drive.google.com/file/d/1_uS955AT6kgAaRkK990PPQEIKHyNBT3S/view?usp=sharing}

\video[P.F.Ac]{rolloutsacrobot}{https://drive.google.com/file/d/1QVQ563fvoLqzvNX4PllIibzX-GHoiw16/view?usp=sharing}

\video[P.F.S6]{rolloutsswimmer}{https://drive.google.com/file/d/1ez6sqCTZmHdSnf86lK6NEsnzyi-O25-B/view?usp=sharing}

\video[P.F.S6(D)]{rolloutsswimmerdpgdata}{https://drive.google.com/file/d/1Ge2_lHVKfgTl_xvqJiGXShcB9iKnopW1/view?usp=sharing}

\video[P.F.SN]{rolloutsmultipleswimmer}{https://drive.google.com/file/d/1BFVbTMliHln4urxrSQJUk6uELqsWSPzx/view?usp=sharing}
\video[P.F.SN(Z)]{rolloutsmultipleswimmerzeroshot}{https://drive.google.com/file/d/15dEUgf5T4ddehMgZiVQ2FtXGqJi9JIDn/view?usp=sharing}

\video[P.F.Ch]{rolloutscheetah}{https://drive.google.com/file/d/14_pWw7sj_RspHqieRK59dGhp_GeZmk1W/view?usp=sharing}

\video[P.F.Wa]{rolloutswalker}{https://drive.google.com/file/d/1xlQssmDDEmo6RUiGldttT0YUTjZ875G6/view?usp=sharing}

\video[P.F.JA]{rolloutsjaco}{https://drive.google.com/file/d/1gGELN8rha56ryGoItoTinWP4sJcOju78/view?usp=sharing}

\video[P.F.MS]{rolloutsmultiplesystems}{https://drive.google.com/file/d/1ApANiv6BF3OniGyUocWO_jhwvJpYwG7B/view?usp=sharing}

\video[P.F.MC]{rolloutsmultiplesystemswithcheetah}{https://drive.google.com/file/d/1agUC9P-cTRUuLyDQexN6ilA87Y_TvtEq/view?usp=sharing}

\video[P.F.JR]{rolloutsrealjaco}{https://drive.google.com/file/d/1o9rF-IHhhAldRYjJF6unBHslI9KbAFI7/view?usp=sharing}

\video[P.P.Pe]{rolloutsparametrizedpendulum}{https://drive.google.com/file/d/1RDYmNVvpNOFboE0V0FuIVJKxPpoCaU4u/view?usp=sharing}

\video[P.P.Ca]{rolloutsparametrizedcartpole}{https://drive.google.com/file/d/1nExfmnLtJjjLWZbBI0T_OfbOykMgtjtU/view?usp=sharing}

\video[P.P.S6]{rolloutsparametrizedswimmer}{https://drive.google.com/file/d/1DKS0pbdY4GWsnoqC6B5IC3CGGR7BcKxY/view?usp=sharing}

\video[P.P.Ch]{rolloutsparametrizedcheetah}{https://drive.google.com/file/d/1AbHjrBFDrGbDKLlPKcsvRBmZFtSlftME/view?usp=sharing}

\video[P.P.JA]{rolloutsparametrizedjaco}{https://drive.google.com/file/d/1JbFRQMVcugMDoYTm2cfRPCVUZg-H6G9h/view?usp=sharing}

\video[P.P.MS]{rolloutsparametrizedmultiplesystems}{https://drive.google.com/file/d/1HZ8E3oIj7F-9DgWDLDkaT0D_S_eFAOgi/view?usp=sharing}

\video[P.I.Pe]{rolloutssysidpendulum}{https://drive.google.com/file/d/1BGIGUt1JjKVXB9t4TyTh6h61IfZ68dVK/view?usp=sharing}

\video[P.I.Ca]{rolloutssysidcartpole}{https://drive.google.com/file/d/19U2nyNxHlq8gG5dLi-NZgjZAJLrOqS0l/view?usp=sharing}

\video[P.I.S6]{rolloutssysidswimmer}{https://drive.google.com/file/d/14SgcWIx9mEWbT6aepNA4sOFluMkHfb5E/view?usp=sharing}

\video[P.I.Ch]{rolloutssysidcheetah}{https://drive.google.com/file/d/1Zn6ct7rbIqOcvfz9y5ADRQbQM46f5TDu/view?usp=sharing}

\video[P.I.JA]{rolloutssysidjaco}{https://drive.google.com/file/d/1O-9AwA_7MHH4fFmdNjWcIH5w68l_KTz8/view?usp=sharing}

\video[C.F.Pe]{controlpendulum}{https://drive.google.com/file/d/1FtJh3n6tsCl6n3G4JOPBNzxOpH0nkbhD/view?usp=sharing}

\video[C.F.Ca]{controlcartpole}{https://drive.google.com/file/d/1cTqgsB5R_Aez_hVhKvFp1uSZZcA5eqOa/view?usp=sharing}

\video[C.F.Ac]{controlacrobot}{https://drive.google.com/file/d/10YS8_MQVg0VETvZp1JrW3wc-zDEFTx16/view?usp=sharing}

\video[C.F.S6]{controlswimmer}{https://drive.google.com/file/d/1ML7Wv0WiUtVwWFKNrUhTKa2rgKQxDwq2/view?usp=sharing}

\video[C.F.SN]{controlmultipleswimmer}{https://drive.google.com/file/d/11VyrLzhp1Rg_CES4LcnJ2gr8CozKYyPw/view?usp=sharing}

\video[C.F.SN(b)]{controlmultipleswimmerbaseline}{https://drive.google.com/file/d/13jcIjdju63Gv4ZVTAHQS9KMS9PjDWiaE/view?usp=sharing}

\video[C.F.Ch(m)]{controlcheetah}{https://drive.google.com/file/d/1F_G5izSPvgehhmd3PGaLPc0smbGLWA_e/view?usp=sharing}

\video[C.F.Ch(k)]{controlcheetahmultiplerewards}{https://drive.google.com/file/d/1oXJFJjMxjf0HQtcosloa_09FXczv4cTj/view?usp=sharing}

\video[C.F.Wa(k)]{controlwalkermultiplerewards}{https://drive.google.com/file/d/1XSX16-ctwBx74d4aMd6_-QiU1O1Iwt0K/view?usp=sharing}

\video[C.F.JA(o)]{controljacofullpose}{https://drive.google.com/file/d/1-R4qC7NHacsDlkjhnwr3w4Gf_89AfCQw/view?usp=sharing}

\video[C.F.JA(a)]{controljacopalmpose}{https://drive.google.com/file/d/1nbb-yobuX2sFyAWJM_TvOI3usc6DbVVo/view?usp=sharing}

\video[C.F.MS]{controlmultiplesystems}{https://drive.google.com/file/d/1xZme1bxvUWQeb9fWFelECR7YKUXur4XU/view?usp=sharing}

\video[C.P.Pe]{controlparametrizedpendulum}{https://drive.google.com/file/d/1feluwsqTE79SwQTsoDO23fW4iBBY3DdT/view?usp=sharing}

\video[C.P.Ca]{controlparametrizedcartpole}{https://drive.google.com/file/d/1rLZBnKX0nuqpnfy4wD8PFLp-hpu8kUuD/view?usp=sharing}

\video[C.P.SN]{controlparametrizedswimmern}{https://drive.google.com/file/d/1ym-2X7-C-MzrxHPw1CU_etH3xJsiSlOo/view?usp=sharing}

\video[C.P.Ch]{controlparametrizedcheetah}{https://drive.google.com/file/d/18CDuW3kGG535vES3aRbklWr58AcUdqKE/view?usp=sharing}

\video[C.P.JA(o)]{controlparametrizedjacofullpose}{https://drive.google.com/file/d/13itwFAwmKIwpX4Iebu5eFuBvmNs8byYg/view?usp=sharing}

\video[C.P.JA(a)]{controlparametrizedjacopalmpose}{https://drive.google.com/file/d/1v70z_aVnCdkYXEyI3Ldo45FBSS5Dzluh/view?usp=sharing}

\video[C.P.MS]{controlparametrizedmultiplesystems}{https://drive.google.com/file/d/1kHeYkekArneNqrrbPT0rVHsyAzG1wezr/view?usp=sharing}

\video[C.I.Pe]{controlidentifiedpendulum}{https://drive.google.com/file/d/1dwMHYwLBqLP4nqhMLBePTMu-7SA9yICc/view?usp=sharing}

\video[C.I.Ca]{controlidentifiedcartpole}{https://drive.google.com/file/d/1ZZQe5QujXLlWxvF8t-HasG_PxakxU5Zw/view?usp=sharing}

\video[C.I.SN]{controlidentifiedswimmern}{https://drive.google.com/file/d/1FiVVN4GRUlYFvcfYqLARho47yT9ahxOg/view?usp=sharing}

\video[C.I.Ch]{controlidentifiedcheetah}{https://drive.google.com/file/d/1oMxkGoEfsBKza8--q4tiVz9MQAETw-8O/view?usp=sharing}

\video[C.I.JA(o)]{controlidentifiedjacofullpose}{https://drive.google.com/file/d/16CXPR_FF1nMx5wiYiaMPOkTLR0G-s-h3/view?usp=sharing}

\video[C.I.JA(a)]{controlidentifiedjacopalmpose}{https://drive.google.com/file/d/1LKPkK709HU4V3iTgTJaYuFb5XKDBIlax/view?usp=sharing}

\begin{document}

\twocolumn[
\icmltitle{\papertitle}

\icmlsetsymbol{equal}{*}

\begin{icmlauthorlist}
\icmlauthor{Alvaro Sanchez-Gonzalez}{dm}
\icmlauthor{Nicolas Heess}{dm}
\icmlauthor{Jost Tobias Springenberg}{dm}
\icmlauthor{Josh Merel}{dm}
\icmlauthor{Martin Riedmiller}{dm}
\icmlauthor{Raia Hadsell}{dm}
\icmlauthor{Peter Battaglia}{dm}
\end{icmlauthorlist}

\icmlaffiliation{dm}{DeepMind, London, United Kingdom}

\icmlcorrespondingauthor{Alvaro Sanchez-Gonzalez}{alvarosg@google.com}
\icmlcorrespondingauthor{Peter Battaglia}{peterbattaglia@google.com}

\icmlkeywords{Machine Learning, ICML}

\vskip 0.3in
]

\printAffiliationsAndNotice{}  

\begin{abstract}

Understanding and interacting with everyday physical scenes requires rich knowledge about the structure of the world, represented either implicitly in a value or policy function, or explicitly in a transition model. Here we introduce a new class of learnable models---based on \emph{graph networks}---which implement an inductive bias for object- and relation-centric representations of complex, dynamical systems. Our results show that as a forward model, our approach supports accurate predictions from real and simulated data, and surprisingly strong and efficient generalization, across eight distinct physical systems which we varied parametrically and structurally. We also found that our inference model can perform system identification. Our models are also differentiable, and support online planning via gradient-based trajectory optimization, as well as offline policy optimization. Our framework offers new opportunities for harnessing and exploiting rich knowledge about the world, and takes a key step toward building machines with more human-like representations of the world.
\end{abstract}

\section{Introduction}
\setcounter{footnote}{1}

\begin{figure}[t]
\begin{center}
\centerline{\includegraphics[width=\columnwidth]{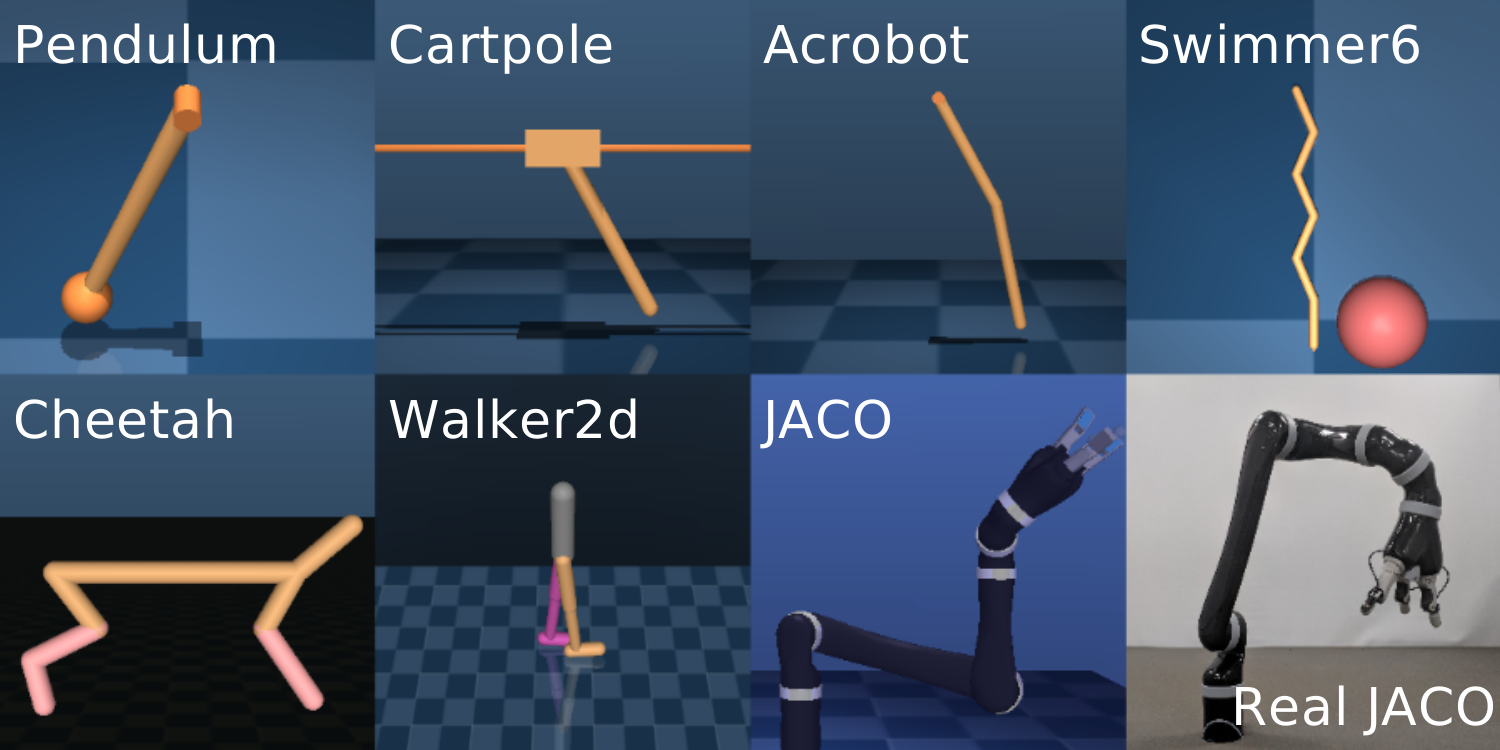}}\vspace{0.2cm}\centerline{\includegraphics[width=\columnwidth]{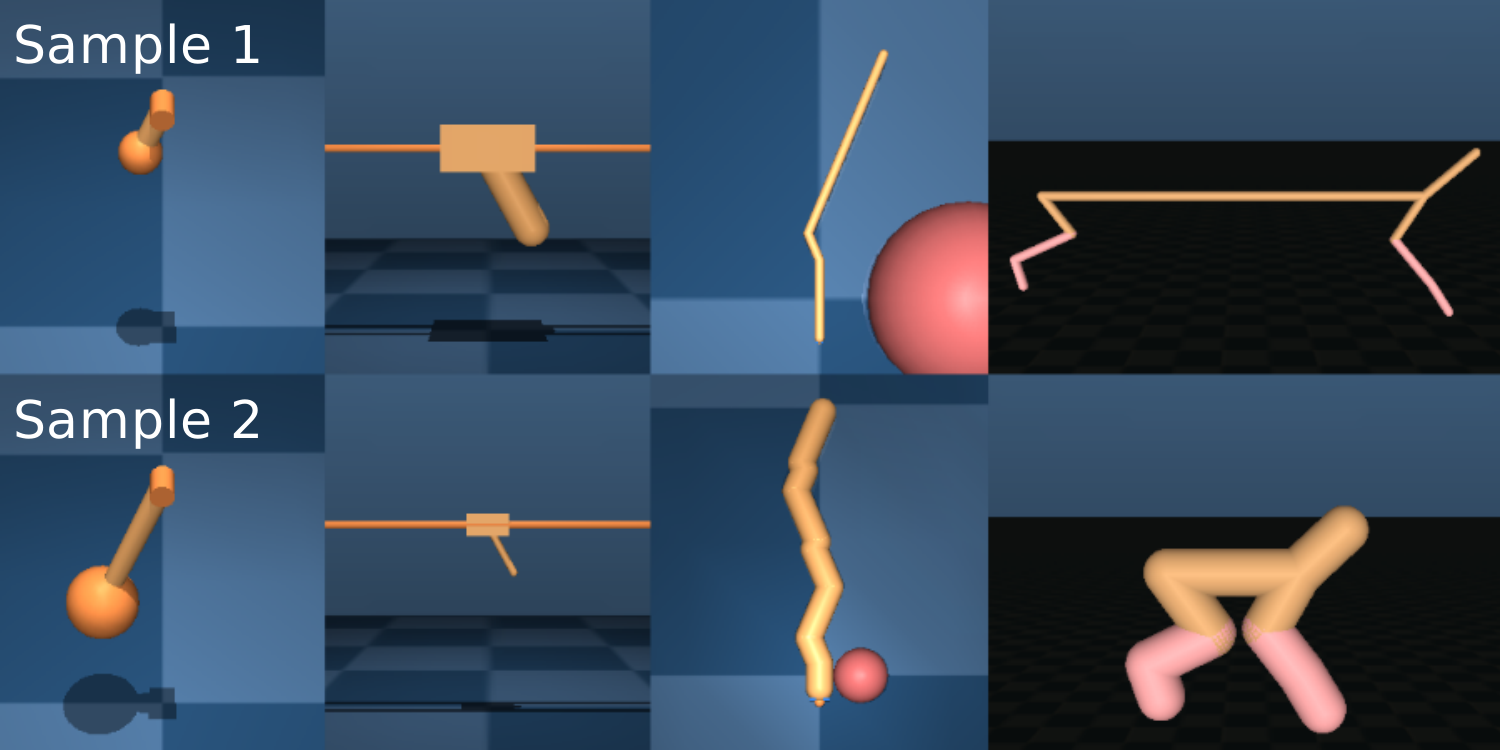}}
\vspace{-0.1in}
\caption{(Top) Our experimental physical systems. (Bottom) Samples of parametrized versions of these systems (see videos: \href{\environmentrandomtrajectories}{link}). 
}
\label{fig:environments}
\end{center}
\vskip -0.2in
\end{figure}

Many domains, such as mathematics, language, and physical systems, are combinatorially complex. The possibilities scale rapidly with the number of elements. For example, a multi-link chain can assume shapes that are exponential in the number of angles each link can take, and a box full of bouncing balls yields trajectories which are exponential in the number of bounces that occur. How can an intelligent agent understand and control such complex systems?

A powerful approach is to represent these systems in terms of objects\footnote{``Object'' here refers to entities generally, rather than physical objects exclusively.} and their relations, applying the same object-wise computations to all objects, and the same relation-wise computations to all interactions. This allows for combinatorial generalization to scenarios never before experienced, whose underlying components and compositional rules are well-understood. For example, particle-based physics engines make the assumption that bodies follow the same dynamics, and interact with each other following similar rules, e.g., via forces, which is how they can simulate limitless scenarios given different initial conditions.

Here we introduce a new approach for learning and controlling complex systems, by implementing a structural inductive bias for object- and relation-centric representations.
Our approach uses ``graph networks'' (GNs), a class of neural networks that can learn functions on graphs  \cite{scarselli2009graph,li2015gated,battaglia2016interaction,gilmer2017neural}. 
In a physical system, the GN lets us represent the bodies (objects) with the graph's nodes and the joints (relations) with its edges. During learning, knowledge about body dynamics is encoded in the GN's node update function, interaction dynamics are encoded in the edge update function, and global system properties are encoded in the global update function. Learned knowledge is shared across the elements of the system, which supports generalization to new systems composed of the same types of body and joint building blocks. 

Across seven complex, simulated physical systems, and one real robotic system (see Figure~\ref{fig:environments}), our experimental results show that our GN-based \emph{forward models} support accurate and generalizable predictions, \emph{inference models}\footnote{We use the term ``inference'' in the sense of ``abductive inference''---roughly, constructing explanations for (possibly partial) observations---and not probabilistic inference, per se.} support system identification in which hidden properties are abduced from observations, and \emph{control algorithms} yield competitive performance against strong baselines. This work represents the first general-purpose, learnable physics engine that can handle complex, 3D physical systems. Unlike classic physics engines, our model has no specific a priori knowledge of physical laws, but instead leverages its object- and relation-centric inductive bias to learn to approximate them via supervised training on current-state/next-state pairs. 

Our work makes three technical contributions: GN-based forward models, inference models, and control algorithms. The forward and inference models are based on treating physical systems as graphs and learning about them using GNs. Our control algorithm uses our forward and inference models for planning and policy learning.

(For full algorithm, implementation, and methodological details, as well as videos from all of our experiments, please see the Supplementary~Material.)

\section{Related Work}

Our work draws on several lines of previous research.
Cognitive scientists have long pointed to rich generative models as central to perception, reasoning, and decision-making \cite{craik1967nature,johnson1980mental,miall1996forward,spelke2007core,battaglia2013simulation}.
Our core model implementation is based on the broader class of graph neural networks (GNNs) \cite{scarselli2005graph,scarselli2009computational,scarselli2009graph,bruna2013spectral,li2015gated,henaff2015deep,duvenaud2015convolutional,dai2016discriminative,defferrard2016convolutional,niepert2016learning,kipf2016semi,battaglia2016interaction,watters2017visual,raposo2017discovering,santoro2017simple,bronstein2017geometric,gilmer2017neural}.
One of our key contributions is an approach for learning physical dynamics models \cite{grzeszczuk1998neuroanimator,fragkiadaki2015learning,battaglia2016interaction,chang2016compositional,watters2017visual,ehrhardt2017learning,amos2018learning}.
Our inference model shares similar aims as approaches for learning system identification explicitly \cite{yu2017preparing,peng2017sim2real}, learning policies that are robust to hidden property variations \cite{rajeswaran2016eopt}, and learning exploration strategies in uncertain settings \cite{schmidhuber1991curious,sun2011planning,houthooft2016curiosity}.
We use our learned models for model-based planning in a similar spirit to classic approaches which use pre-defined models \cite{li2004iterative,tassa2008receding,tassa2014control}, and our work also relates to learning-based approaches for model-based control \cite{atkeson1997comparison,deisenroth2011pilco,levine2014learning}.
We also explore jointly learning a model and policy \cite{heess2015learning,gu2016continuous,nagabandi2017neural}.
Notable recent, concurrent work \cite{wang2018nervenet} used a GNN to approximate a policy, which complements our use of a related architecture to approximate forward and inference models.

\section{Model}
\label{sec:model}

\begin{figure*}[t]
\vskip 0.05in
\centerline{\includegraphics[width=\textwidth]{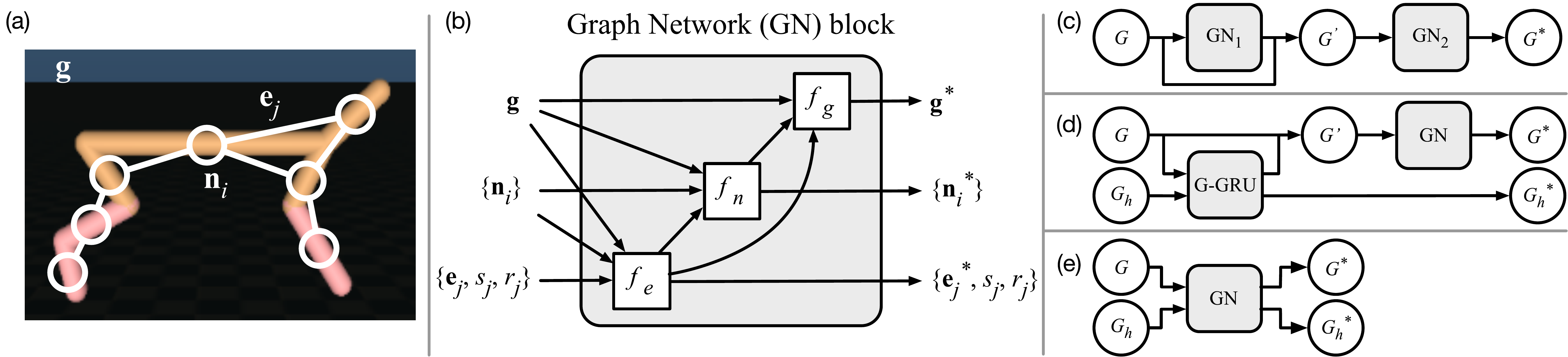}}
\vspace{-0.07in}
\caption{Graph representations and GN-based models. (a) A physical system's bodies and joints can be represented by a graph's nodes and edges, respectively. (b) A GN block takes a graph as input and returns a graph with the same structure but different edge, node, and global features as output (see Algorithm~\ref{alg:graph_network}). (c) A feed-forward GN-based forward model for learning one-step predictions. (d) A recurrent GN-based forward model. (e) A recurrent GN-based inference model for system identification.}
\label{fig:schematic}
\vskip -0.125in
\end{figure*}

\paragraph{Graph representation of a physical system.}
Our approach is founded on the idea of representing physical systems as graphs: the bodies and joints correspond to the nodes and edges, respectively, as depicted in Figure~\ref{fig:schematic}a. 
Here a (directed) \emph{graph} is defined as ${G=\left(\mathbf{g}, \{\mathbf{n}_i\}_{i=1\cdots N_n},\{\mathbf{e}_j,s_j,r_j\}_{j=1\cdots N_e}\right)}$, where $\textbf{g}$ is a vector of global features, $\{\mathbf{n}_i\}_{i=1\cdots N_n}$ is a set of nodes where each $\mathbf{n}_i$ is a vector of node features, and $\{\mathbf{e}_j, s_j, r_j\}_{j=1\cdots N_e}$ is a set of directed edges where $\mathbf{e}_j$ is a vector of edge features, and $s_j$ and $r_j$ are the indices of the sender and receiver nodes, respectively.

We distinguish between static and dynamic properties in a physical scene, which we represent in separate graphs. A static graph $G_s$ contains static information about the parameters of the system, including global parameters (such as the time step, viscosity, gravity, etc.), per body/node parameters (such as mass, inertia tensor, etc.), and per joint/edge parameters (such as joint type and properties, motor type and properties, etc.).
A dynamic graph $G_d$ contains information about the instantaneous state of the system. This includes each body/node's 3D Cartesian position, 4D quaternion orientation, 3D linear velocity, and 3D angular velocity.\footnote{Some physics engines, such as Mujoco \cite{todorov2012mujoco}, represent systems using ``generalized coordinates'', which sparsely encode degrees of freedom rather than full body states. Generalized coordinates offer advantages such as preventing bodies connected by joints from dislocating (because there is no degree of freedom for such displacement). In our approach, however, such representations do not admit sharing as naturally because there are different input and output representations for a body depending on the system's constraints.}
Additionally, it contains the magnitude of the actions applied to the different joints in the corresponding edges.

\paragraph{Graph networks.}
The GN architectures introduced here generalize interaction networks (IN) \cite{battaglia2016interaction} in several ways. They include global representations and outputs for the state of a system, as well as per-edge outputs. They are defined as ``graph2graph'' modules (i.e., they map input graphs to output graphs with different edge, node, and global features), which can be composed in deep and recurrent neural network (RNN) configurations. A core GN block (Figure~\ref{fig:schematic}b) contains three sub-functions---edge-wise,~$f_e$, node-wise,~$f_n$, and global,~$f_g$---which can be implemented using standard neural networks. Here we use multi-layer perceptrons (MLP). A single feedforward GN pass can be viewed as one step of message-passing on a graph \cite{gilmer2017neural}, where $f_e$ is first applied to update all edges, $f_n$ is then applied to update all nodes, and $f_g$ is finally applied to update the global feature. See Algorithm~\ref{alg:graph_network} for details. 

\begin{algorithm}[t!]
\caption{Graph network, GN}
  \label{alg:graph_network}
\begin{algorithmic}
    \STATE {\bfseries Input:} Graph, $G = (\textbf{g}, \{\textbf{n}_i\}, \{\textbf{e}_j, s_j, r_j\})$ 
  \FOR{each edge $\{\textbf{e}_j, s_j, r_j\}$}
  \STATE Gather sender and receiver nodes $\textbf{n}_{s_j}, \textbf{n}_{r_j}$
  \STATE Compute output edges, 
  $\textbf{e}^*_j = f_{e}(\textbf{g}, \textbf{n}_{s_j}, \textbf{n}_{r_j}, \textbf{e}_j)$
  \ENDFOR
  \FOR{each node $\{\textbf{n}_i\}$}
     
     \STATE Aggregate $\textbf{e}^*_j$ per receiver, $\hat{\textbf{e}}_{i} = \sum_{j / r_j=i} \textbf{e}^*_j$
     \STATE Compute node-wise features,
  $\textbf{n}^*_i =  f_{n}(\textbf{g}, \textbf{n}_i, \hat{\textbf{e}}_{i})$
  \ENDFOR
  \STATE Aggregate all edges and nodes $\hat{\textbf{e}} = \sum_{j} \textbf{e}^*_j$, $\hat{\textbf{n}} = \sum_{i} \textbf{n}^*_i$
  \STATE Compute global features,
  $\textbf{g}^* =  f_{g}(\textbf{g}, \hat{\textbf{n}}, \hat{\textbf{e}})$
  \STATE {\bfseries Output:} Graph, $G^* = (\textbf{g}^*, \{\textbf{n}^*_i\}, \{\textbf{e}^*_j, s_j, r_j\})$ 
\end{algorithmic}
\end{algorithm}

\begin{figure*}[t]
\vskip 0.05in
\begin{center}
\centerline{\includegraphics[width=0.9\columnwidth]{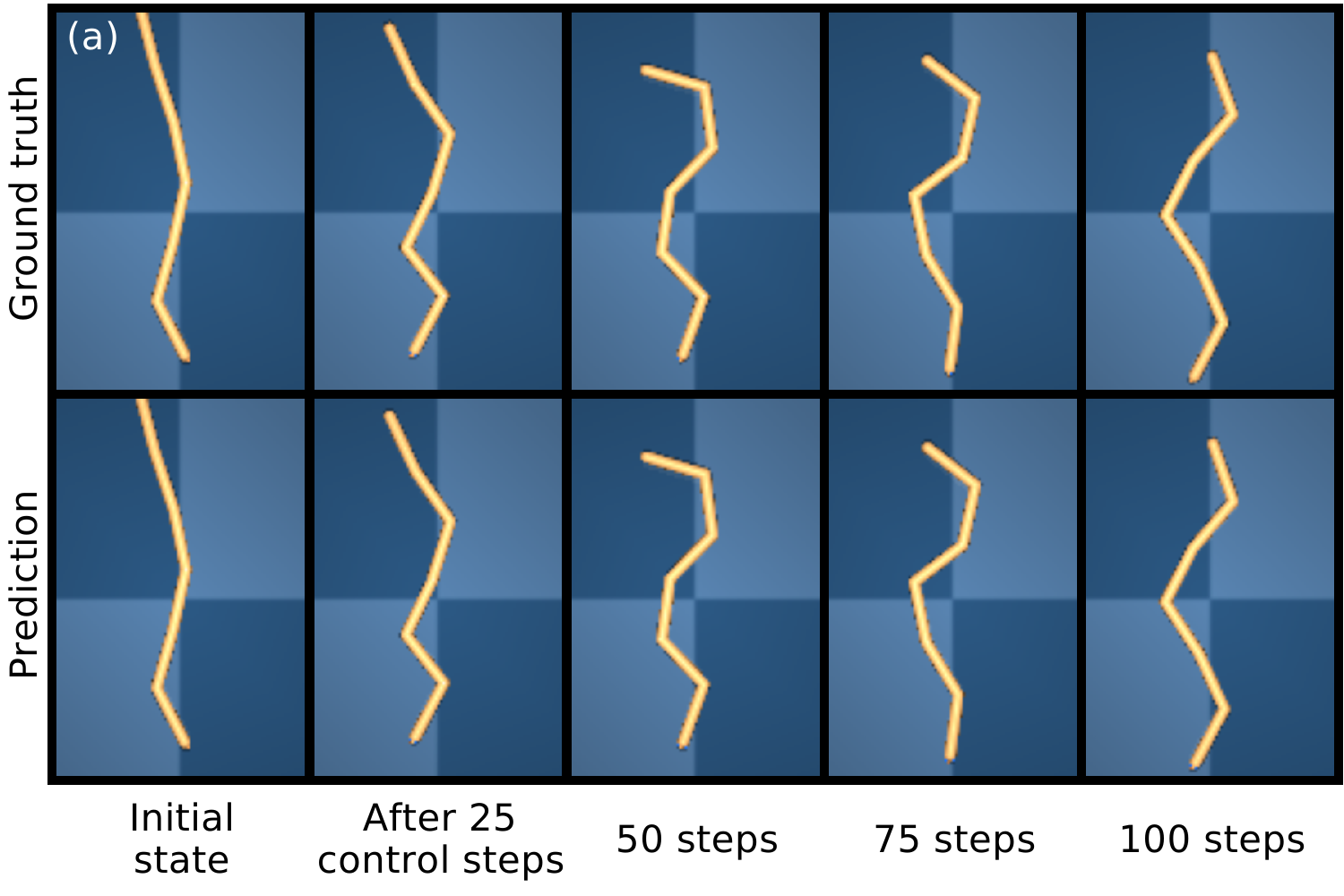}\hspace{0.2cm}
\includegraphics[width=0.92\columnwidth]{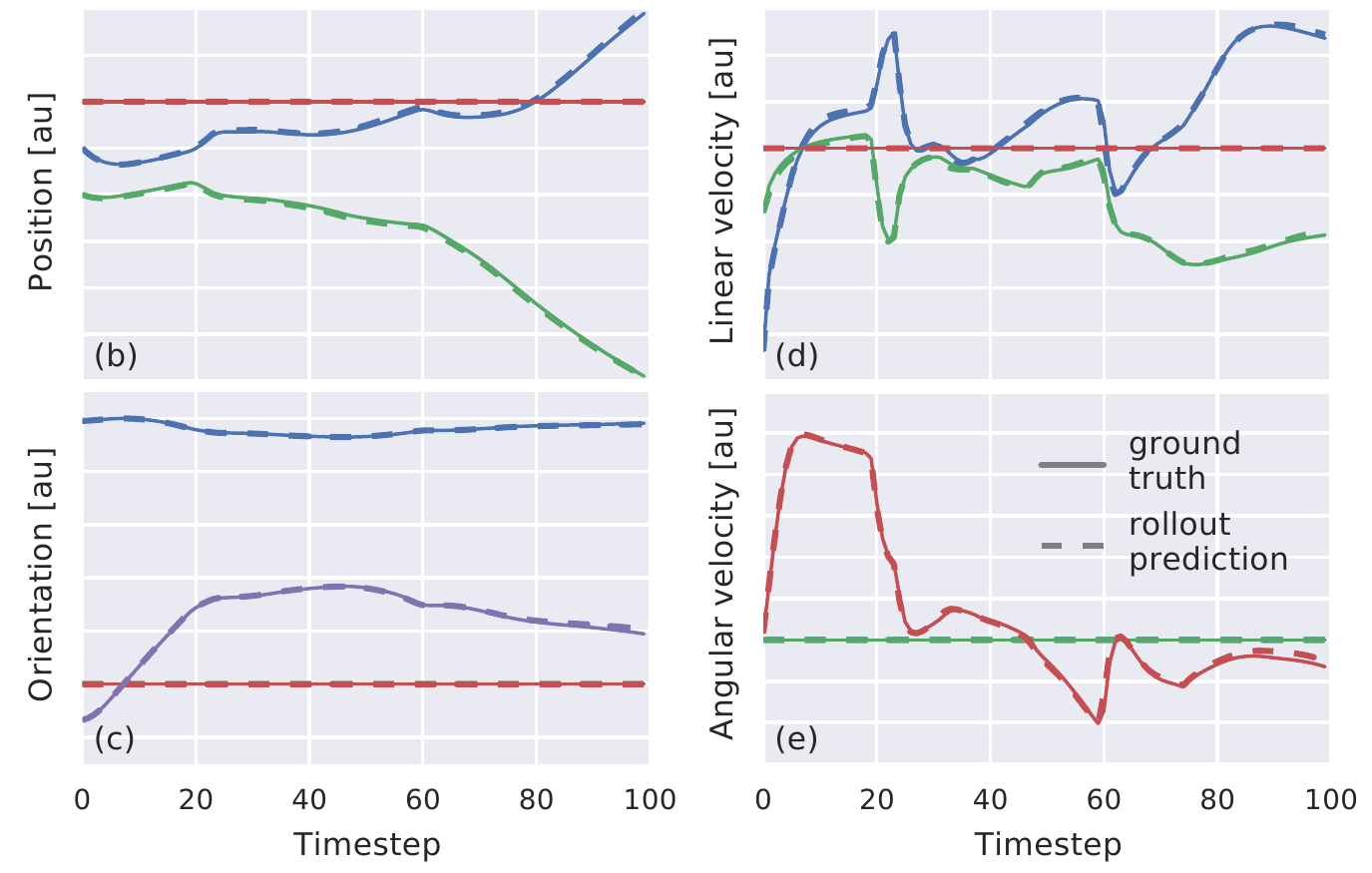}}
\vspace{-0.05in}
\caption{Evaluation rollout in a Swimmer6.
Trajectory videos are here: \href{\rolloutsswimmer}{\rolloutsswimmername}. 
(a) Frames of ground truth and predicted states over a 100 step trajectory. (b-e) State sequence predictions for link \#3 of the Swimmer. The subplots are (b) $x,y,z$-position, (c) $q0,q1,q2,q3$-quaternion orientation, (d) $x,y,z$-linear velocity, and (e) $x,y,z$-angular velocity. [au] indicates arbitrary units.}
\label{fig:swimmer_rollout}
\end{center}
\vskip -0.3in
\end{figure*}

\paragraph{Forward models.}
For prediction, we introduce a GN-based forward model for learning to predict future states from current ones. 
It operates on one time-step, and contains two GNs composed sequentially in a ``deep'' arrangement (unshared parameters; see Figure~\ref{fig:schematic}c). The first GN takes an input graph, $G$, and produces a latent graph, $G'$. This $G'$ is concatenated\footnote{We define the term ``graph-concatenation'' as combining two graphs by concatenating their respective edge, node, and global features. We define ``graph-splitting'' as splitting the edge, node, and global features of one graph to form two new graphs with the same structure.\label{fn:graph-cat-split}} with $G$ (e.g., a graph skip connection), and provided as input to the second GN, which returns an output graph, $G^*$. Our forward model training optimizes the GN so that $G^*$'s $\{\mathbf{n}_i\}$ features reflect predictions about the states of each body across a time-step. The reason we used two GNs was to allow all nodes and edges to communicate with each other through the $\mathbf{g}'$ output from the first GN. Preliminary tests suggested this provided large performance advantages over a single IN/GN (see ablation study in SM~Figure~\ref{sm:fig:ablation_study}).

We also introduce a second, recurrent GN-based forward model, which contains three RNN sub-modules (GRUs, \cite{cho2014properties}) applied across all edges, nodes, and global features, respectively, before being composed with a GN block (see Figure~\ref{fig:schematic}d). 

Our forward models were all trained to predict state differences, so to compute absolute state predictions we updated the input state with the predicted state difference. To generate a long-range \emph{rollout} trajectory, we repeatedly fed absolute state predictions and externally specified control inputs back into the model as input, iteratively. As data pre- and post-processing steps, we normalized the inputs and outputs to the GN model.

\paragraph{Inference models.}
System identification refers to inferences about unobserved properties of a dynamic system based on its observed behavior. It is important for controlling systems whose unobserved properties influence the control dynamics. Here we consider ``implicit'' system identification, in which inferences about unobserved properties are not estimated explicitly, but are expressed in latent representations which are made available to other mechanisms.

We introduce a recurrent GN-based inference model, which observes only the dynamic states of a trajectory and constructs a latent representation of the unobserved, static properties (i.e., performs implicit system identification). %
It takes as input a sequence of dynamic state graphs, $G_d$, under some control inputs, and returns an output, $G^*(T)$, after $T$ time steps. This $G^*(T)$ is then passed to a one-step forward model by graph-concatenating it with an input dynamic graph, $G_d$. 
The recurrent core takes as input, $G_d$, and hidden graph, $G_h$, which are graph-concatenated\textsuperscript{\ref{fn:graph-cat-split}} and passed to a GN block (see Figure~\ref{fig:schematic}e). The graph returned by the GN block is graph-split\textsuperscript{\ref{fn:graph-cat-split}} to form an output, $G^*$, and updated hidden graph, $G_h^*$.
The full architecture can be trained jointly, and learns to infer unobserved properties of the system from how the system's observed features behave, and use them to make more accurate predictions. 

\paragraph{Control algorithms.}
For control, we exploit the fact that the GN is differentiable to use our learned forward and inference models for model-based planning within a classic, gradient-based trajectory optimization regime, also known as model-predictive control (MPC). We also develop an agent which simultaneously learns a GN-based model and policy function via Stochastic Value Gradients (SVG) \cite{heess2015learning}.
\footnote{MPC and SVG are deeply connected: in MPC the control inputs are optimized given the initial conditions in a single episode, while in SVG a policy function that maps states to controls is optimized over states experienced during training.}

\section{Methods}
\label{sec:methods}
\paragraph{Environments.}
Our experiments involved seven actuated Mujoco simulation environments (Figure~\ref{fig:environments}). Six were from the ``DeepMind Control Suite'' \cite{tassa2018deepmind}---Pendulum, Cartpole, Acrobot, Swimmer, Cheetah, Walker2d---and one was a model of a JACO commercial robotic arm. We generated training data for our forward models by applying simulated random controls to the systems, and recording the state transitions. We also trained models from recorded trajectories of a real JACO robotic under human control during a stacking task.

In experiments that examined generalization and system identification, we created a dataset of versions of several of our systems---Pendulum, Cartpole, Swimmer, Cheetah and JACO--- with procedurally varied parameters and structure. We varied continuous properties such as link lengths, body masses, and motor gears. In addition, we also varied the number of links in the Swimmer's structure, from 3-15 (we refer to a swimmer with $N$ links as Swimmer$N$).

\paragraph{MPC planning.}
We used our GN-based forward model to implement MPC planning by maximizing a dynamic-state-dependent reward along a trajectory from a given initial state. We used our GN forward model to predict the $N$-step trajectories ($N$ is the planning \emph{horizon}) induced by proposed action sequences, as well as the total reward associated with the trajectory. We optimized these action sequences by backpropagating gradients of the total reward with respect to the actions, and minimizing the negative reward by gradient descent, iteratively.

\begin{figure}[t]
\vskip 0.05in
\begin{center}
\centerline{\includegraphics[width=\columnwidth]{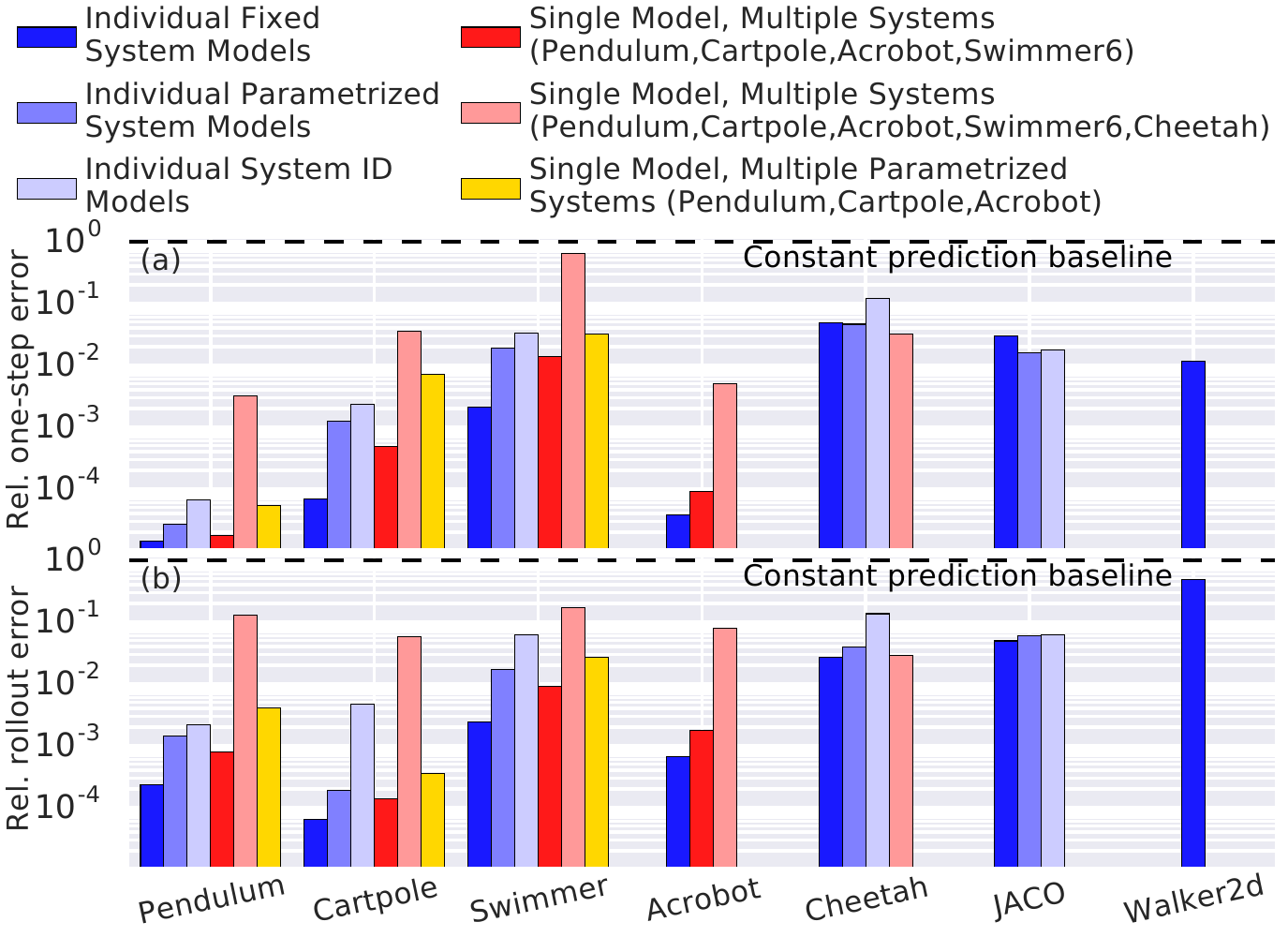}}
\vspace{-0.1in}
\caption{(a) One-step and (b) 100-step rollout errors for different models and training (different bars) on different test data (x-axis labels), relative to the constant prediction baseline (black dashed line). Blue bars are GN models trained on single systems. Red and yellow bars are GN models trained on multiple systems, with (yellow) and without (red) parametric variation. Note that including Cheetah in multiple system training caused performance to diminish (light red vs dark red bars), which suggests sharing might not always be beneficial.
}
\label{fig:quantitative_comparison}
\end{center}
\vskip -0.2in
\end{figure}
\begin{figure}[t]
\vskip 0.05in
\begin{center}
\centerline{\includegraphics[width=\columnwidth]{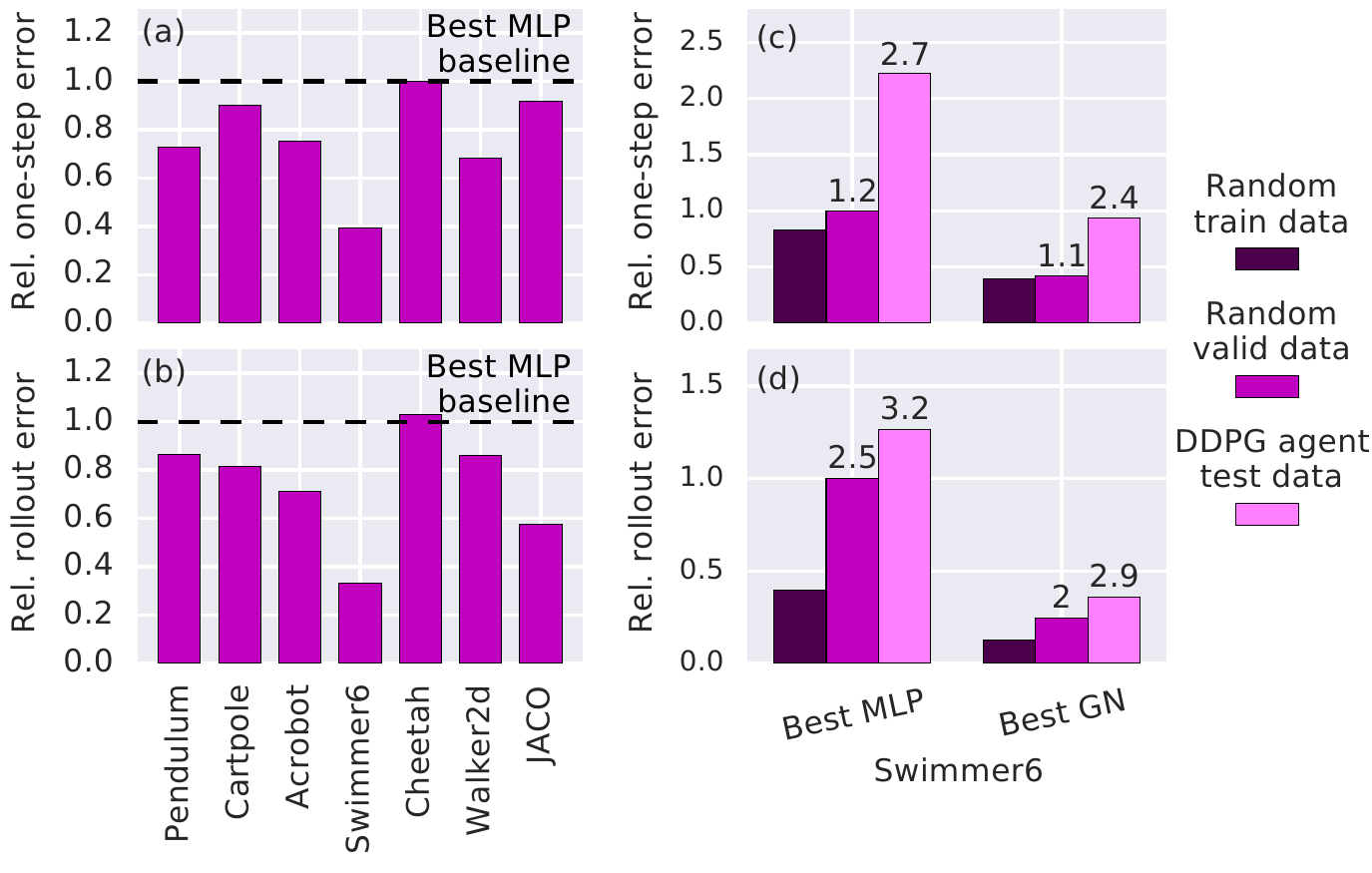}}
\vspace{-0.1in}
\caption{Prediction errors, on (a) one-step and (b) 20-step evaluations, between the best MLP baseline and the best GN model after 72 hours of training.
Swimmer6 prediction errors, on (c) one-step and (d) 20-step evaluations, between the best MLP baseline and the best GN model for data in the training set (dark), data in the validation set (medium), and data from DDPG agent trajectories (light). The numbers above the bars indicate the ratio between the corresponding generalization test error (medium or light) and the training error (dark).
}
\label{fig:mlp_comparison}
\end{center}
\vskip -0.2in
\end{figure}

\paragraph{Model-based reinforcement learning.}
To investigate whether our GN-based model can benefit reinforcement learning (RL) algorithms, we used our model within an SVG regime \citep{heess2015learning}. The GN forward model was used as a differentiable environment simulator to obtain a gradient of the expected return (predicted based on the next state generated by a GN) with respect to a parameterized, stochastic policy, which was trained jointly with the GN. For our experiments we used a single step prediction (SVG(1)) and compared to sample-efficient model-free RL baselines using either stochastic policies (SVG(0)) or deterministic policies via the Deep Deterministic Policy Gradients (DDPG) algorithm \citep{Lillicrap16} (which is also used as a baseline in the MPC experiments).

\paragraph{Baseline comparisons.}
As a simple baseline, we compared our forward models' predictions to a \emph{constant prediction baseline}, which copied the input state as the output state. 
We also compared our GN-based forward model with a learned, MLP baseline, which we trained to make forward predictions using the same data as the GN model. We replaced the core GN with an MLP, and flattened and concatenated the graph-structured GN input and target data into a vector suitable for input to the MLP. We swept over 20 unique hyperparameter combinations for the MLP architecture, with up to 9 hidden layers and 512 hidden nodes per layer.

As an MPC baseline, with a pre-specified physical model, we used a Differential Dynamic Programming algorithm \citep{tassa2008receding, tassa2014control} that had access to the ground-truth Mujoco model. We also used the two model-free RL agents mentioned above, SVG(0) and DDPG, as baselines in some tests. Some of the trajectories from a DDPG agent in Swimmer6 were also used to evaluate generalization of the forward models.

\paragraph{Prediction performance evaluation.}
Unless otherwise specified, we evaluated our models on squared one-step dynamic state differences (\emph{one-step error}) and squared trajectory differences (\emph{rollout error}) between the prediction and the ground truth.
We calculated independent errors for position, orientation, linear velocity angular velocity, and normalized them individually to the constant prediction baseline. After normalization, the errors were averaged together. All errors reported are calculated for 1000 100-step sequences from the test set.

\section{Results: Prediction}
\label{sec:results_prediction}

\paragraph{Learning a forward model for a single system.}
Our results show that the GN-based model can be trained to make very accurate forward predictions under random control.
For example, the ground truth and model-predicted trajectories for Swimmer6 were both visually and quantitatively indistinguishable (see Figure~\ref{fig:swimmer_rollout}). 
Figure~\ref{fig:quantitative_comparison}'s black bars show that the predictions across most other systems were far better than the constant prediction baseline.
As a stronger baseline comparison, Figures~\ref{fig:mlp_comparison}a-b show that our GN model had lower error than the MLP-based model in 6 of the 7 simulated control systems we tested. This was especially pronounced for systems with much repeated structure, such as the Swimmer, while for systems with little repeated structure, such as Pendulum, there was negligible difference between the GN and MLP baseline. These results suggest that a GN-based forward model is very effective at learning predictive dynamics in a diverse range of complex physical systems.

We also found that the GN generalized better than the MLP baseline from training to test data, as well as across different action distributions. 
Figures~\ref{fig:mlp_comparison}c-d show that for Swimmer6, the relative increase in error from training to test data, and to data recorded from a learned DDPG agent, was smaller for the GN model than for the MLP baseline.
We speculate that the GN's superior generalization is a result of implicit regularization due to its inductive bias for sharing parameters across all bodies and joints; the MLP, in principle, could devote disjoint subsets of its computations to each body and joint, which might impair generalization.

\begin{figure}[t!]
\vskip 0.05in
\begin{center}
\centerline{\includegraphics[width=1\columnwidth]{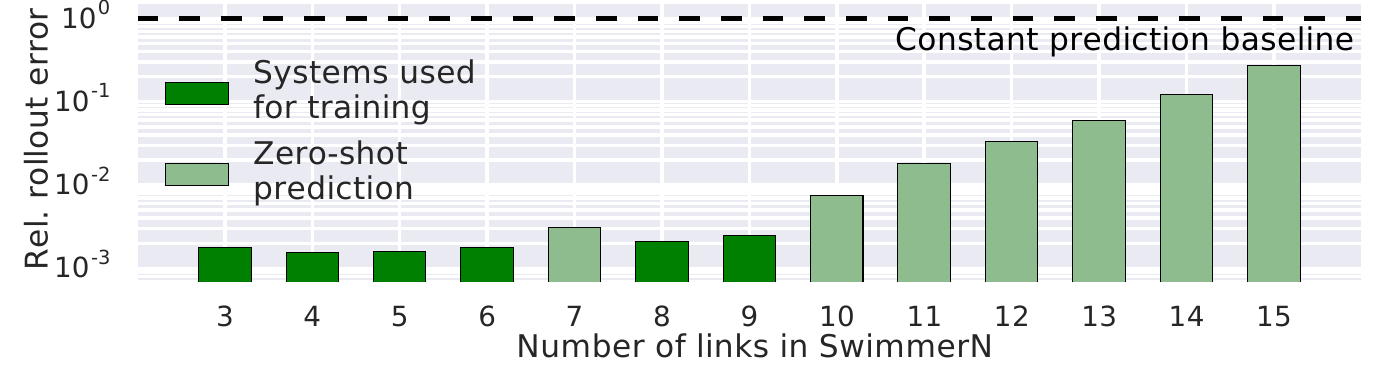}}
\vspace{-0.1in}
\caption{Zero-shot dynamics prediction. The bars show the 100-step rollout error of a model trained on a mixture of 3-6 and 8-9 link Swimmers, and tested on Swimmers with 3-15 links. The dark bars indicate test Swimmers whose number of links the model was trained on (video: \href{\rolloutsmultipleswimmer}{\rolloutsmultipleswimmername}), the light bars indicate Swimmers it was not trained on (video: \href{\rolloutsmultipleswimmerzeroshot}{\rolloutsmultipleswimmerzeroshotname}).
}
\label{fig:multiple_swimmer_rollout_baseline}
\end{center}
\vskip -0.2in
\end{figure}

\paragraph{Learning a forward model for multiple systems.}
Another important feature of our GN model is that it is very flexible, able to handle wide variation across a system's properties, and across systems with different structure.
We tested how it learned forward dynamics of systems with continuously varying static parameters, using a new dataset where the underlying systems' bodies and joints had different masses, body lengths, joint angles, etc. These static state features were provided to the model via the input graphs' node and edge attributes.
Figure~\ref{fig:quantitative_comparison} shows that the GN model's forward predictions were again accurate, which suggests it can learn well even when the underlying system properties vary.

We next explored the GN's inductive bias for body- and joint-centric learning by testing whether a single model can make predictions across multiple systems that vary in their number of bodies and the joint structure. Figure~\ref{fig:multiple_swimmer_rollout_baseline} shows that when trained on a mixed dataset of Swimmers with 3-6, 8-9 links, the GN model again learned to make accurate forward predictions.  
We pushed this even further by training a single GN model on multiple systems, with completely different structures, and found similarly positive results (see Figure~\ref{fig:quantitative_comparison}, red and yellow bars). 
This highlights a key difference, in terms of general applicability, between GN and MLP models: the GN can naturally operate on variably structured inputs, while the MLP requires fixed-size inputs.

The GN model can even generalize, zero-shot, to systems whose structure was held out during training, as long as they are composed of bodies and joints similar to those seen during training. For the GN model trained on Swimmers with 3-6, 8-9 links, we tested on held-out Swimmers with 7 and 10-15 links. Figure~\ref{fig:multiple_swimmer_rollout_baseline} shows that zero-shot generalization performance is very accurate for 7 and 10 link Swimmers, and degrades gradually from 11-15 links. Still, their trajectories are visually very close to the ground truth (video:  \href{\rolloutsmultipleswimmerzeroshot}{\rolloutsmultipleswimmerzeroshotname}).

\begin{figure}[t!]
\begin{center}
\centerline{\includegraphics[width=1\columnwidth]{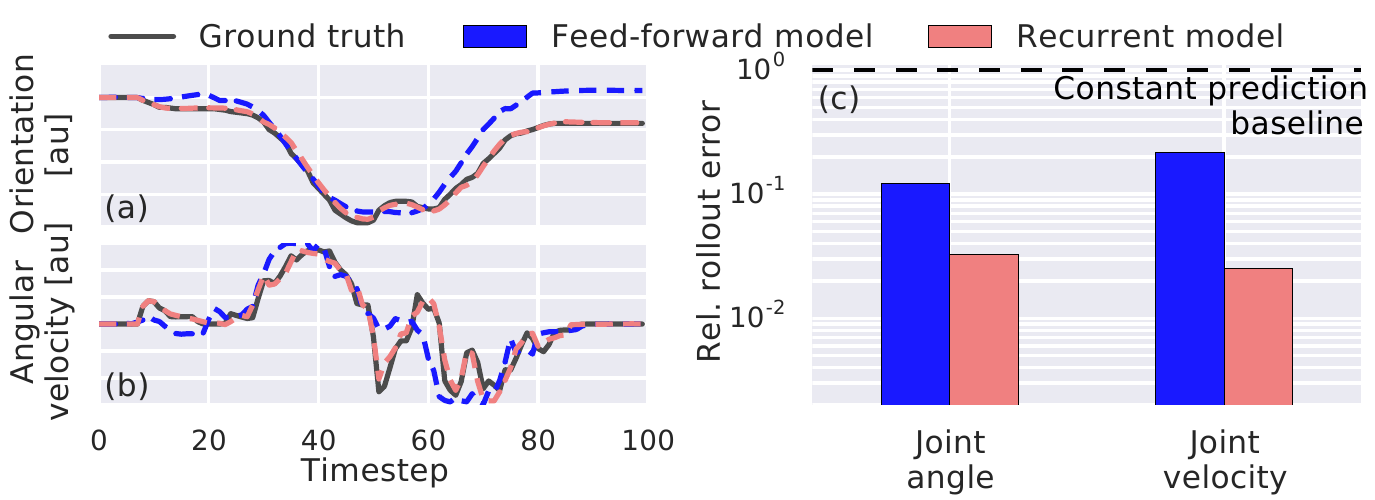}}
\vspace{-0.1in}
\caption{Real and predicted test trajectories of a JACO robot arm. The recurrent model tracks the ground truth (a) orientations and (b) angular velocities closely. (c) The total 100-step rollout error was much better for the recurrent model, though the feed-forward model was still well below the constant prediction baseline. A video of a Mujoco rendering of the true and predicted trajectories: \href{\rolloutsrealjaco}{\rolloutsrealjaconame}.
}
\label{fig:real_jaco_rollout}
\end{center}
\vskip -0.25in
\end{figure}

\paragraph{Real robot data.}
To evaluate our approach's applicability to the real world, we trained GN-based forward models on real JACO proprioceptive data; under manual control by a human performing a stacking task.
We found the feed-forward GN performance was not as accurate as the recurrent GN forward model\footnote{This might result from lag or hysteresis which induces long-range temporal dependencies that the feed-forward model cannot capture.}: Figure~\ref{fig:real_jaco_rollout} shows a representative predicted trajectory from the test set, as well as overall performance. These results suggest that our GN-based forward model is a promising approach for learning models in real systems.

\section{Results: Inference}
\begin{figure}[t!]
\vskip 0.05in
\begin{center}
\centerline{\includegraphics[width=\columnwidth]{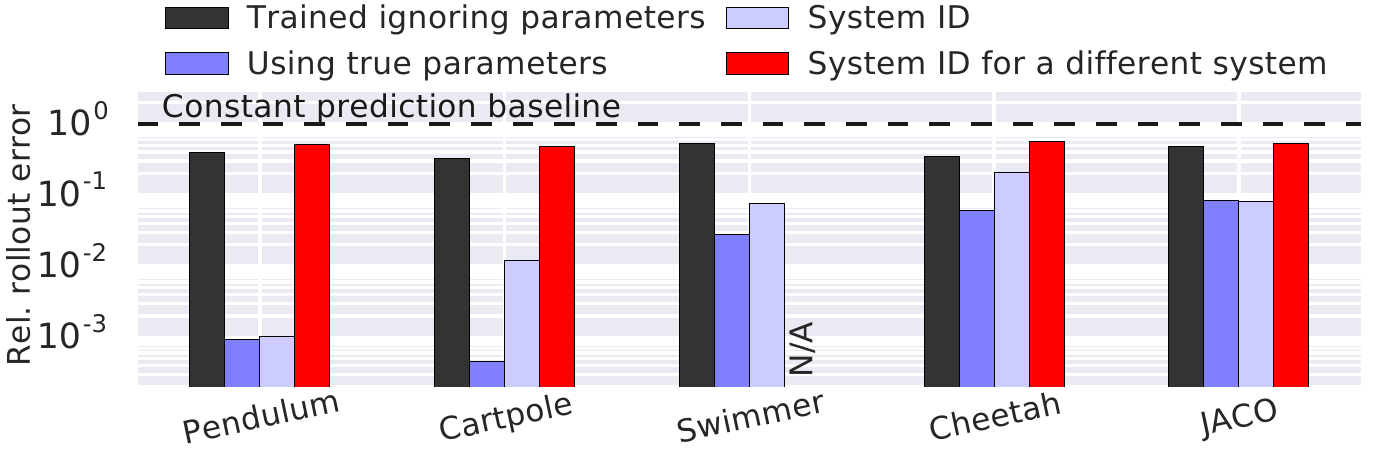}}
\vspace{-0.1in}
\caption{System identification performance. The y-axis represents 100-step rollout error, relative to the trivial constant prediction baseline (black dashed line). The baseline GN-based model (black bars) with no system identification module performs worst. A model which was always provided the true static parameters (medium blue bars) and thus did not require system identification performed best. A model without explicit access to the true static parameters, but with a system identification module (light blue bars), performed generally well, sometimes very close to the model which observed the true parameters. But when that same model was presented with an ID phase whose hidden parameters were different (but from the same distribution) from its prediction phase (red bars), its performance was similar or worse than the model (black) with no ID information available. (The N/A column is because our Swimmer experiments always varied the number of links as well as parameters, which meant the inferred static graph could not be concatenated with the initial dynamic graph.)}
\label{fig:quantitative_comparison_sysid}
\end{center}
\vskip -0.2in
\end{figure}

In many real-world settings the system's state is partially observable. Robot arms often use joint angle and velocity sensors, but other properties such as mass, joint stiffness, etc. are often not directly measurable. 
We applied our system identification inference model (see Model~Section~\ref{sec:model}) to a setting where only the dynamic state variables (i.e., position, orientation, and linear and angular velocities) were observed, and found it could support accurate forward predictions (during its ``prediction phase'') after observing randomly controlled system dynamics during an initial 20-step ``ID phase'' (see Figure~\ref{fig:quantitative_comparison_sysid}).

\begin{figure}[t!]
\begin{center}
\centerline{\includegraphics[width=\columnwidth]{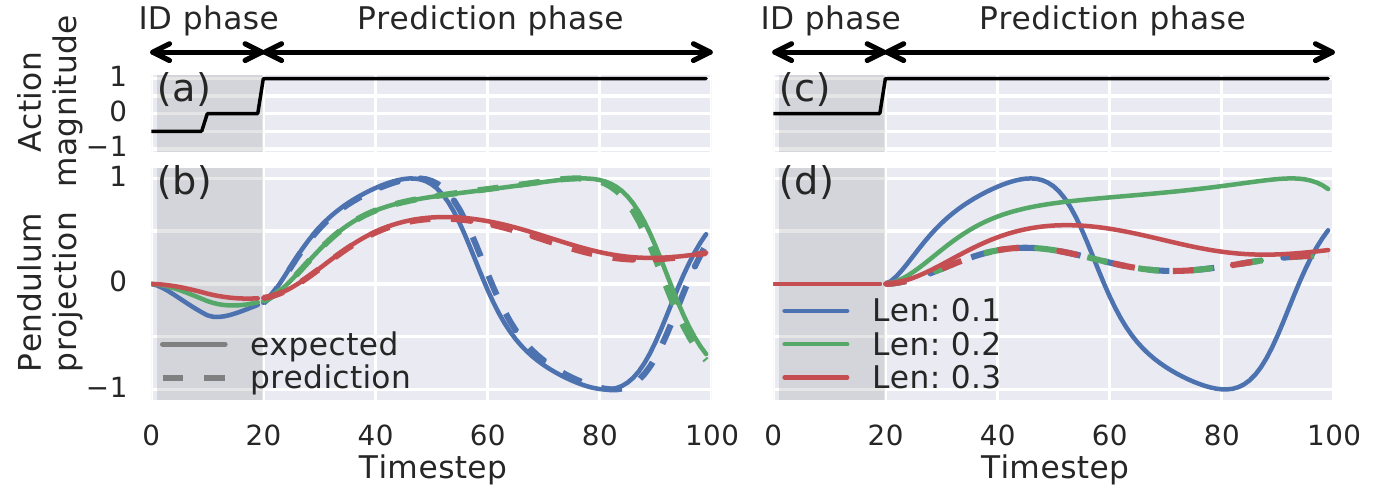}}
\vspace{-0.1in}
\caption{System identification analysis in Pendulum. (a) Control inputs are applied to three Pendulums with different, unobservable lengths during the 20-step ID phase, which makes the system identifiable. (b) The model's predicted trajectories (dashed curves) track the ground truth (solid curves) closely in the subsequent 80-step prediction phase. (c) No control inputs are applied to the same systems during the ID phase, which makes the system identifiable. (d) The model's predicted trajectories across systems are very different from the ground truth.
}
\label{fig:sysid_pendulum}
\end{center}
\vskip -0.35in
\end{figure}

To further explore the role of our GN-based system identification, we contrasted the model's predictions after an ID phase, which contained useful control inputs, against an ID phase that did not apply control inputs, across three different Pendulum systems with variable, unobserved lengths.
Figure~\ref{fig:sysid_pendulum} shows that the GN forward model with an identifiable ID phase makes very accurate predictions, but with an unidentifiable ID phase its predictions are very poor. 

A key advantage of our system ID approach is that once the ID phase has been performed for some system, the inferred representation can be stored and reused to make trajectory predictions from different initial states of the system. 
This contrasts with an approach that would use an RNN to both infer the system properties and use them throughout the trajectory, which thus would require identifying the same system from data each time a new trajectory needs to be predicted given different initial conditions.

\begin{figure}[ht]
\vskip 0.0in
\begin{center}
\centerline{\includegraphics[width=\columnwidth]{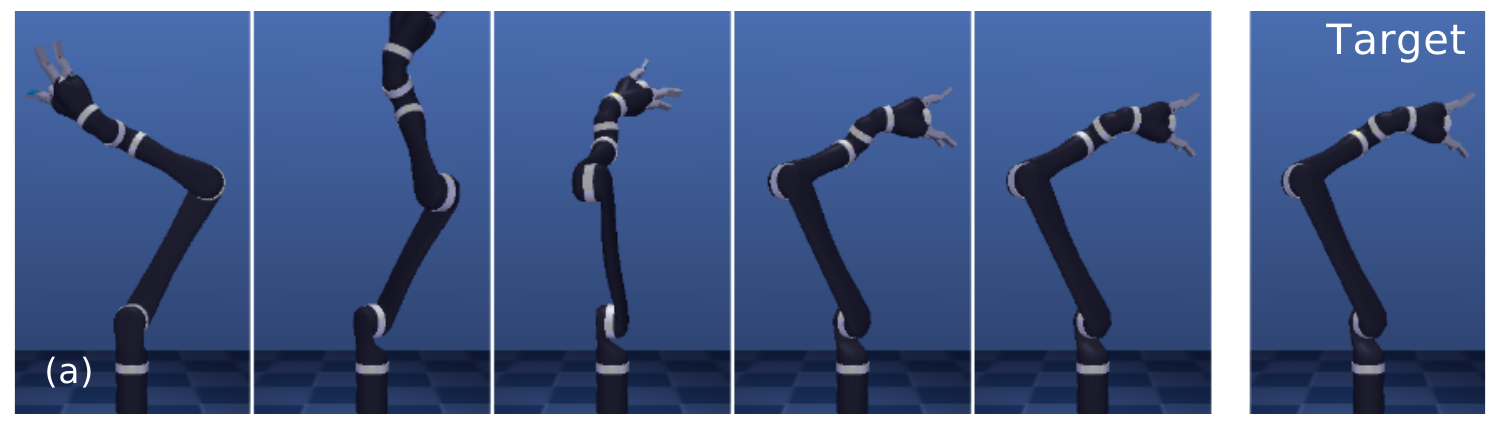}}
\centerline{\includegraphics[width=\columnwidth]{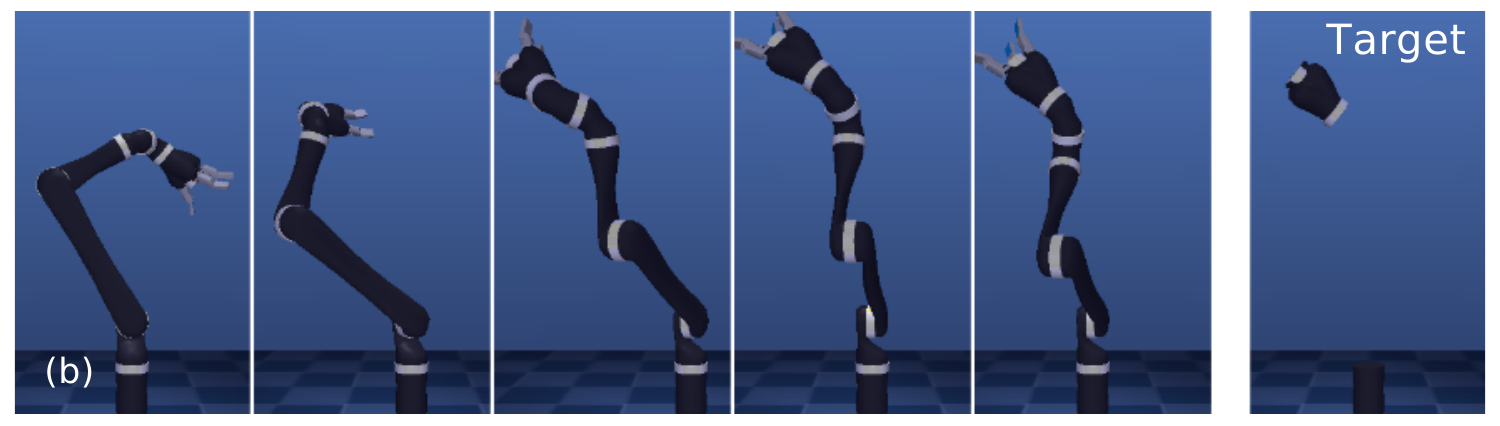}}
\vspace{-0.05in}
\caption{Frames from a 40-step GN-based MPC trajectory of the simulated JACO arm. (a) Imitation of the pose of each individual body of the arm (13 variables x 9 bodies). (b) Imitation of only the palm's pose (13 variables). The full videos are here: \href{\controljacofullpose}{\controljacofullposename} and \href{\controljacopalmpose}{\controljacopalmposename}.}
\label{fig:jaco_poses}
\end{center}
\vskip -0.3in
\end{figure}

\section{Results: Control}

Differentiable models can be valuable for model-based sequential decision-making, and here we explored two approaches for exploiting our GN model in continuous control. 

\paragraph{Model-predictive control for single systems.}
We trained a GN forward model and used it for MPC by optimizing the control inputs via gradient descent to maximize predicted reward under a known reward function. 
We found our GN-based MPC could support planning in all of our control systems, across a range of reward functions. For example, Figure~\ref{fig:jaco_poses} shows frames of simulated JACO trajectories matching a target pose and target palm location, respectively, under MPC with a 20-step planning horizon.

\begin{figure}[ht]
\vskip 0.05in
\begin{center}
\centerline{\includegraphics[width=\columnwidth]{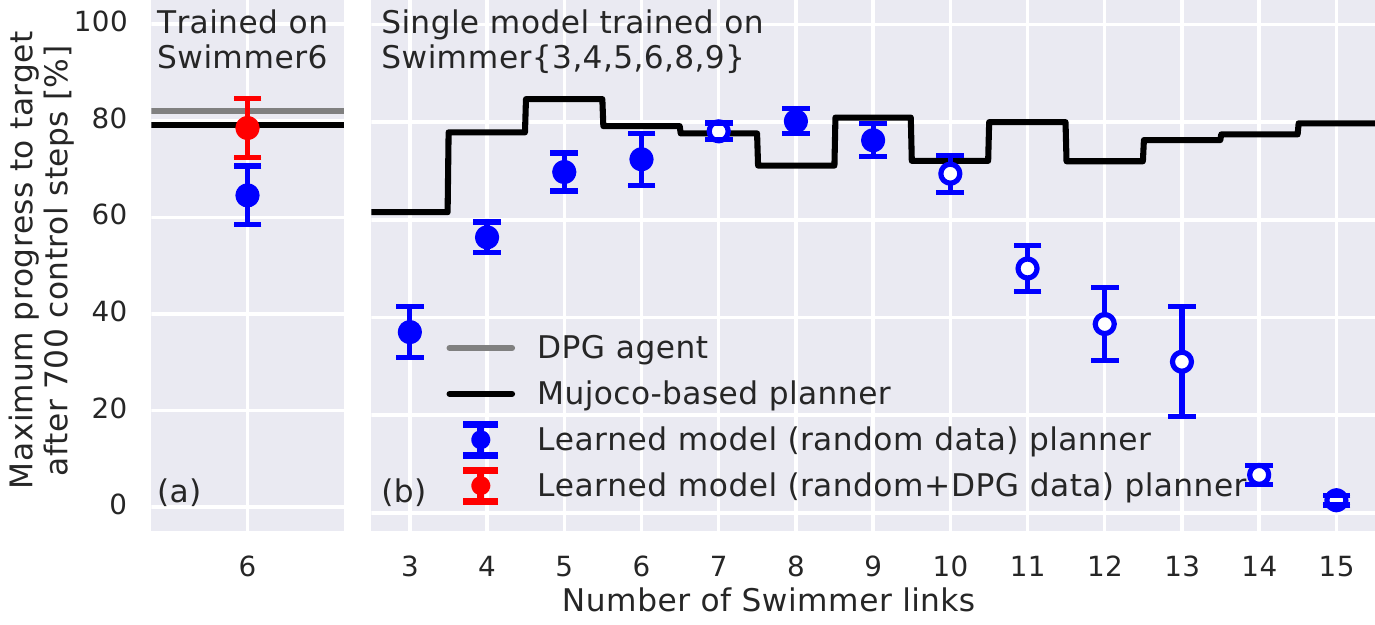}}
\vspace{-0.1in}
\caption{GN-based MPC performance (\% distance to target after 700 steps) for (a) model trained on Swimmer6 and (b) model trained on Swimmers with 3-15 links (see Figure~\ref{fig:multiple_swimmer_rollout_baseline}). In (a), GN-based MPC (blue point) is almost as good as the Mujoco-based planner (black line) and trained DDPG (grey line) baselines. When the GN-based MPC's model is trained on a mixture of random and DDPG agent Swimmer6 trajectories (red point), its performance is as good as the strong baselines. In (b) the GN-based MPC (blue point) (video: \href{\controlmultipleswimmer}{\controlmultipleswimmername}) is competitive with a Mujoco-based planner baseline (black) (video: \href{\controlmultipleswimmerbaseline}{\controlmultipleswimmerbaselinename}) for 6-10 links, but is worse for 3-5 and 11-15 links. Note, the model was not trained on the open points, 7 and 10-15 links, which correspond to zero-shot model generalization for control. Error bars indicate mean and standard deviation across 5 experimental runs.
}
\label{fig:control_swimmer_comparison_full}
\end{center}
\vskip -0.2in
\end{figure}

In the Swimmer6 system with a reward function that maximized the head's movement toward a randomly chosen target, GN-based MPC with a 100-step planning horizon selected control inputs that resulted in coordinated, swimming-like movements.
Despite the fact that the Swimmer6 GN model used for MPC was trained to make one-step predictions under random actions, its swimming performance was close to both that of a more sophisticated planning algorithm which used the true Mujoco physics as its model, as well as that of a learned DDPG agent trained on the system (see Figure~\ref{fig:control_swimmer_comparison_full}a). And when we trained the GN model using a mixture of both random actions and DDPG agent trajectories, there was effectively no difference in performance between our approach, versus the Mujoco planner and learned DDPG agent baselines (see video: \href{\controlswimmer}{\controlswimmername}).

For Cheetah with reward functions for maximizing forward movement, maximizing height, maximizing squared vertical speed, and maximizing squared angular speed of the torso, MPC with a 20-step horizon using a GN model resulted in running, jumping, and other reasonable patterns of movements (see video: \href{\controlcheetahmultiplerewards}{\controlcheetahmultiplerewardsname}).

\paragraph{Model-predictive control for multiple systems.}
Similar to how our forward models learned accurate predictions across multiple systems, we also found they could support MPC across multiple systems (in this video, a single model is used for MPC in Pendulum, Cartpole, Acrobot, Swimmer6 and Cheetah: \href{\controlmultiplesystems}{\controlmultiplesystemsname}).
We also found GN-based MPC could support zero-shot generalization in the control setting, for a single GN model trained on Swimmers with 3-6, 8-9 links, and tested on MPC on Swimmers with 7, 10-15 links. Figure~\ref{fig:control_swimmer_comparison_full}b shows that it performed almost as well as the Mujoco baseline for many of the Swimmers.

\paragraph{Model-predictive control with partial observations.}
Because real-world control settings are often partially observable, we used the system identification GN model (see Sections~\ref{sec:model}~and~\ref{sec:results_prediction}) for MPC under partial observations in Pendulum, Cartpole, SwimmerN, Cheetah, and JACO. The model was trained as in the forward prediction experiments, with an ID phase that applied 20 random control inputs to implicitly infer the hidden properties. Our results show that our GN-based forward model with a system identification module is able to control these systems (Cheetah video: \href{\controlidentifiedcheetah}{\controlidentifiedcheetahname}. All videos are in SM~Table~\ref{sm:table:control}).

\begin{figure}[t]
\vskip 0.05in
\begin{center}
\centerline{\includegraphics[width=\columnwidth]{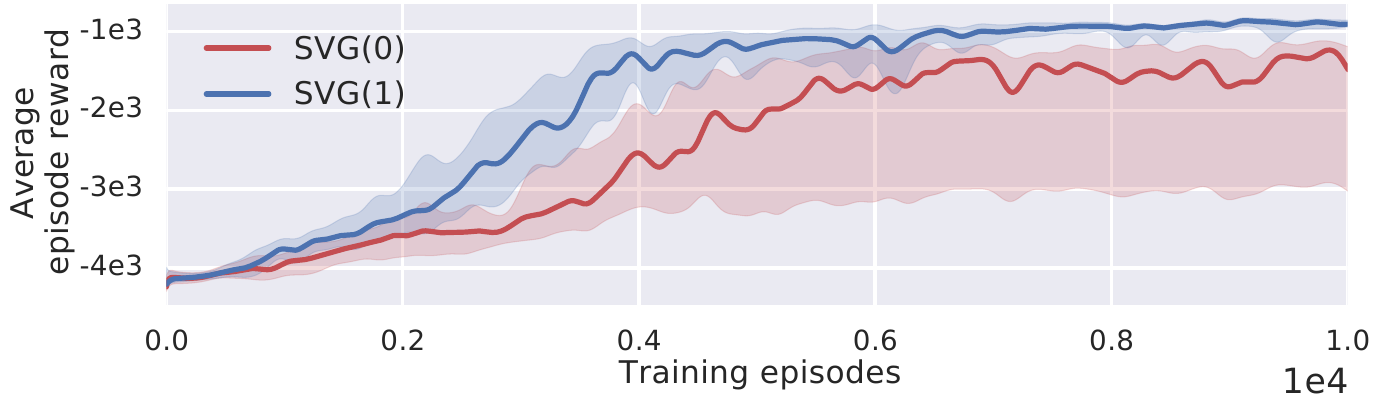}}
\vspace{-0.1in}
\caption{Learning curves for Swimmer6 SVG agents. The GN-based agent (blue) asymptotes earlier, and at a higher performance, than the model-free agent (red). The lines represent median performance for 6 random seeds, with 25 and 75\% confidence intervals.}
\label{fig:model_based_swimmer}
\end{center}
\vskip -0.35in
\end{figure}

\paragraph{Model-based reinforcement learning.}
In our second approach to model-based control, we jointly trained a GN model and a policy function using SVG \cite{heess2015learning}, where the model was used to backpropagate error gradients to the policy in order to optimize its parameters.
Crucially, our SVG agent does not use a pre-trained model, but rather the model and policy were trained simultaneously.\footnote{In preliminary experiments, we found little benefit of pre-training the model, though further exploration is warranted.}
Compared to a model-free agent (SVG(0)), our GN-based SVG agent (SVG(1)) achieved a higher level performance after fewer episodes (Figure~\ref{fig:model_based_swimmer}). For GN-based agents with more than one forward step (SVG(2-4)), however, the performance was not significantly better, and in some cases was worse (SVG(5+)).

\section{Discussion}

This work introduced a new class of learnable forward and inference models, based on ``graph networks'' (GN), which implement an object- and relation-centric inductive bias. Across a range of experiments we found that these models are surprisingly accurate, robust, and generalizable when used for prediction, system identification, and planning in challenging, physical systems.

While our GN-based models were most effective in systems with common structure among bodies and joints (e.g., Swimmers), they were less successful when there was not much opportunity for sharing (e.g., Cheetah). Our approach also does not address a common problem for model-based planners that errors compound over long trajectory predictions.

Some key future directions include using our approach for control in real-world settings, supporting simulation-to-real transfer via pre-training models in simulation, extending our models to handle stochastic environments, and performing system identification over the structure of the system as well as the parameters.
Our approach may also be useful within imagination-based planning frameworks \cite{hamrick2017metacontrol,pascanu2017learning}, as well as integrated architectures with GN-like policies \cite{wang2018nervenet}.

This work takes a key step towards realizing the promise of model-based methods by exploiting compositional representations within a powerful statistical learning framework, and opens new paths for robust, efficient, and general-purpose patterns of reasoning and decision-making.

\bibliography{references}
\bibliographystyle{icml2018}

\appendix
\counterwithin{figure}{section}
\counterwithin{table}{section}
\counterwithin{algorithm}{section}
\onecolumn
\icmltitle{Supplementary Material: \papertitle}

\icmlsetsymbol{equal}{*}

\begin{icmlauthorlist}
\icmlauthor{Alvaro Sanchez-Gonzalez}{dm}
\icmlauthor{Nicolas Heess}{dm}
\icmlauthor{Jost Tobias Springenberg}{dm}
\icmlauthor{Josh Merel}{dm}
\icmlauthor{Martin Riedmiller}{dm}
\icmlauthor{Raia Hadsell}{dm}
\icmlauthor{Peter Battaglia}{dm}
\end{icmlauthorlist}

\icmlaffiliation{dm}{DeepMind, London, United Kingdom}

\vskip 0.3in

\section{Summary of prediction and control videos}
\begin{table}[hb]
\caption{Representative trajectory prediction videos. Each shows several rollouts from different initial states for a single model trained on random control inputs. The labels encode the videos' contents: [\textbf{P}rediction/\textbf{C}ontrol].[\textbf{F}ixed/\textbf{P}arameterized/System~\textbf{I}D].[(System abbreviation)]}
\label{sm:table:rollouts}
\begin{center}
\begin{small}
\begin{tabular}{p{27mm}lll}
\toprule
 & Fixed & Parametrized & System ID \\
\midrule
Pendulum  & 
\href{\rolloutspendulum}{\rolloutspendulumname} & \href{\rolloutsparametrizedpendulum}{\rolloutsparametrizedpendulumname} & 
\href{\rolloutssysidpendulum}{\rolloutssysidpendulumname}\\
Cartpole   & 
\href{\rolloutscartpole}{\rolloutscartpolename} & \href{\rolloutsparametrizedcartpole}{\rolloutsparametrizedcartpolename} & 
\href{\rolloutssysidcartpole}{\rolloutssysidcartpolename}\\
Acrobot   & 
\href{\rolloutsacrobot}{\rolloutsacrobotname}& - & - \\
Swimmer6   & 
\href{\rolloutsswimmer}{\rolloutsswimmername}& - & - \\
\ \ \ (eval. DDPG) & 
\href{\rolloutsswimmerdpgdata}{\rolloutsswimmerdpgdataname}& - & -\\
SwimmerN  & 
\href{\rolloutsmultipleswimmer}{\rolloutsmultipleswimmername}& 
\href{\rolloutsparametrizedswimmer}{\rolloutsparametrizedswimmername} & 
\href{\rolloutssysidswimmer}{\rolloutssysidswimmername}\\
\ \ \ (zero-shot) & 
\href{\rolloutsmultipleswimmerzeroshot}{\rolloutsmultipleswimmerzeroshotname}& -& -\\
Cheetah   & 
\href{\rolloutscheetah}{\rolloutscheetahname} & \href{\rolloutsparametrizedcheetah}{\rolloutsparametrizedcheetahname} & 
\href{\rolloutssysidcheetah}{\rolloutssysidcheetahname}\\
Walker2d & 
\href{\rolloutswalker}{\rolloutswalkername}& - & -\\
JACO & 
\href{\rolloutsjaco}{\rolloutsjaconame}& 
\href{\rolloutsparametrizedjaco}{\rolloutsparametrizedjaconame} & 
\href{\rolloutssysidjaco}{\rolloutssysidjaconame}\\
Multiple systems &  
\href{\rolloutsmultiplesystems}{\rolloutsmultiplesystemsname} &
\href{\rolloutsparametrizedmultiplesystems}{\rolloutsparametrizedmultiplesystemsname}  & - \\
\ \ \ (with cheetah)&  
\href{\rolloutsmultiplesystemswithcheetah}{\rolloutsmultiplesystemswithcheetahname} &
- & - \\
Real JACO  & 
\href{\rolloutsrealjaco}{\rolloutsrealjaconame}& - & -\\
\bottomrule
\end{tabular}
\end{small}
\end{center}
\vskip -0.1in
\end{table}
\begin{table}[hb]
\caption{Representative control trajectory videos. Each shows several MPC trajectories from different initial states for a single trained model. The labels encode the videos' contents: [\textbf{P}rediction/\textbf{C}ontrol].[\textbf{F}ixed/\textbf{P}arameterized/System~\textbf{I}D].[(System abbreviation)]
}
\label{sm:table:control}
\begin{center}
\begin{small}
\begin{tabular}{p{26.6mm}lll}
\toprule
 & Fixed & Parametrized & System ID \\
\midrule
Pendulum (balance)   & \href{\controlpendulum}{\controlpendulumname} & \href{\controlparametrizedpendulum}{\controlparametrizedpendulumname} & \href{\controlidentifiedpendulum}{\controlidentifiedpendulumname}\\
Cartpole (balance)   & \href{\controlcartpole}{\controlcartpolename} & \href{\controlparametrizedcartpole}{\controlparametrizedcartpolename} & \href{\controlidentifiedcartpole}{\controlidentifiedcartpolename}\\
Acrobot (swing up)   & \href{\controlacrobot}{\controlacrobotname} &  - & - \\
Swimmer6 (reach)   & \href{\controlswimmer}{\controlswimmername}  & - & - \\
SwimmerN (reach) & \href{\controlmultipleswimmer}{\controlmultipleswimmername} & 
\href{\controlparametrizedswimmern}{\controlparametrizedswimmernname}& \href{\controlidentifiedswimmern}{\controlidentifiedswimmernname}\\
\ \ \ \ \ \ \ \ \ " \ \ \ \ \ \ \ \ baseline & 
\href{\controlmultipleswimmerbaseline}{\controlmultipleswimmerbaselinename}& -& -\\
Cheetah (move)   & \href{\controlcheetah}{\controlcheetahname} &  \href{\controlparametrizedcheetah}{\controlparametrizedcheetahname} & \href{\controlidentifiedcheetah}{\controlidentifiedcheetahname}\\
Cheetah ($k$ rewards)  & \href{\controlcheetahmultiplerewards}{\controlcheetahmultiplerewardsname} & - & - \\
Walker2d ($k$ rewards)  & \href{\controlwalkermultiplerewards}{\controlwalkermultiplerewardsname} & - & - \\
JACO (imitate pose)    & \href{\controljacofullpose}{\controljacofullposename} &  \href{\controlparametrizedjacofullpose}{\controlparametrizedjacofullposename} & \href{\controlidentifiedjacofullpose}{\controlidentifiedjacofullposename}\\
JACO (imitate palm)    & \href{\controljacopalmpose}{\controljacopalmposename} &  \href{\controlparametrizedjacopalmpose}{\controlparametrizedjacopalmposename}& \href{\controlidentifiedjacopalmpose}{\controlidentifiedjacopalmposename}\\
Multiple systems & \href{\controlmultiplesystems}{\controlmultiplesystemsname} & \href{\controlparametrizedmultiplesystems}{\controlparametrizedmultiplesystemsname} & -\\
\bottomrule
\end{tabular}
\end{small}
\end{center}
\vskip -0.1in
\end{table}

\section{Description of the simulated environments}
\label{sm:sec:environments}

\begin{minipage}{\columnwidth}
\vskip 0.15in
\begin{center}
\begin{small}
\renewcommand{\arraystretch}{1.5}
\begin{tabular}{p{15mm}p{14mm}p{36mm}p{32mm}p{45mm}}
\toprule
 Name \newline (Timestep) & Number \newline of bodies \newline (inc. world) & Generalized \newline coordinates & Actions & Random parametrization\footnote{Density of bodies is kept constant for any changes in size.}  \newline (relative range of variation, \newline uniformly sampled) \\
\midrule
Pendulum \newline (20 ms)  & 2  
& {Total: 1 
\newline 1: angle of pendulum}
& {1: rotation torque at axis}
& Length (0.2-1)
\newline Mass (0.5-3)
\\
Cartpole \newline (10 ms)  & 3
& {Total: 2 
\newline 1: horizontal position of cart
\newline 1: angle of pole}
& {1: horizontal force to cart}
& Mass of cart (0.2-2)
\newline Length of pole (0.3-1) 
\newline Thickness (0.4-2.2) of pole
\\
Acrobot \newline (10 ms)  & 3 
& {Total: 2 
\newline 2: angle of each of the links 
\newline angle of pole}
& {1: rotation force between the links}
& N/A
\\
SwimmerN  \newline (20 ms) & N+1 
& {Total: N+2
\newline 2: 2-d position of head 
\newline 1: angle of head
\newline N-1: angle of rest of links} 
& {N-1: rotation force between the links}
& Number of links (3 to 9 links)
\newline Individual lengths of links (0.3-2)
\newline Thickness (0.5-5)
\\
Cheetah \newline (10 ms)  & 8 
& {Total: 9
\newline 2: 2-d position of torso
\newline 1: angle of torso
\newline 6: thighs, shins and feet angles}
& {6: rotation force at thighs, shins and feet}
& Base angles (-0.1 to 0.1 rad) 
\newline Individual lengths of bodies (0.5-2 approx.)
\newline Thickness (0.5-2)
\\
Walker2d \newline (2.5 ms) & 8
& {Total: 9
\newline 2: 2-d position of torso
\newline 1: angle of torso
\newline 6: thighs, leg and feet angles}
& {6: rotation at hips, knees and ankles}
& N/A
\\
Jaco \newline (100 ms) & 10 
& {Total: 9
\newline 3: angles of coarse joints
\newline 3: angles of fine joints
\newline 3: angles of fingers}
& {9: velocity target at each joint}
& Individual body masses (0.5-1.5) 
\newline Individual motor gears (0.5-1.5).
\\
\bottomrule
\end{tabular}
\end{small}
\end{center}
\vskip -0.1in
\renewcommand{\arraystretch}{1.}
\end{minipage}

\section{System data}

\subsection{Random control}
\label{sm:sec:random_control}

\begin{figure}[ht]
\vskip 0.2in
\begin{center}
\centerline{\includegraphics[width=0.4\columnwidth]{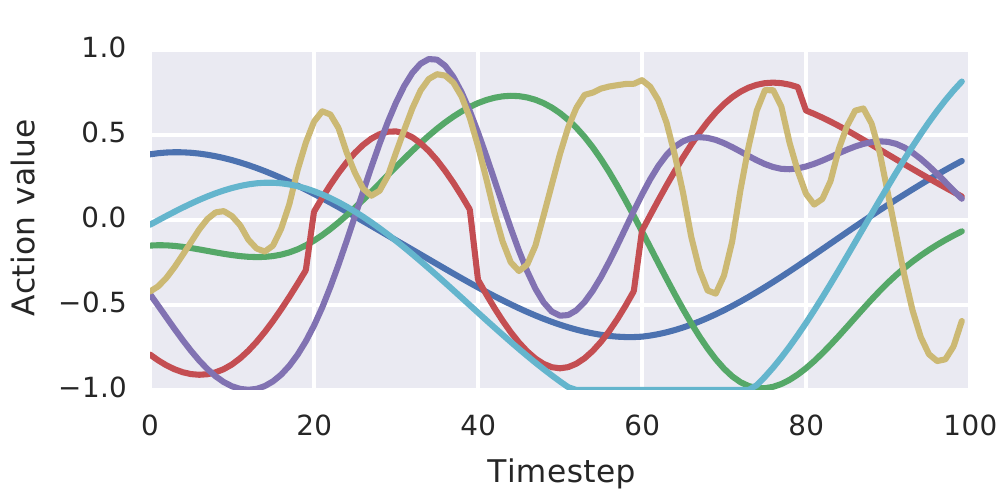}}
\caption{Sample random sequences obtained from the same distribution than that used to generate random system data to train the models. Sample trajectory video:  \href{\environmentrandomtrajectories}{Video}.}
\label{sm:fig:random_actions}
\end{center}
\vskip -0.2in
\end{figure}

Unless otherwise indicated, we applied random control inputs to the system to generate the training data. The control sequences were randomly selected time steps from spline interpolations of randomly generated values (see SM~Figure~\ref{sm:fig:random_actions}). A video of the resulting random system trajectories is here: \href{\environmentrandomtrajectories}{Video}.

\subsection{Datasets}
\label{sm:sec:data}
For each of the individual fixed systems, we generated 10000 100-step sequences corresponding to about $10^6$ supervised training examples. Additionally, we generated 1000 sequences for validation, and 1000 sequences for testing purposes.

In the case of the parametrized environments, we generated 20000 100-step sequences corresponding to about $2\cdot10^6$ supervised training examples. Additionally, we generated 5000 sequences for validation, and 5000 sequences for testing purposes.

Models trained on multiple environments made use of the corresponding datasets mixed within each batch in equal proportion.

\subsection{Real JACO}
\label{sm:sec:real_jaco}

The real JACO data was obtained under human control during a stacking task. It consisted of 2000 (train:1800, valid:100, test:100) 100-step (timestep 40 ms) trajectories. The instantaneous state of the system was represented in this case by proprioceptive information consisting of joint angles (cosine and sine) and joint velocities for each connected body in the JACO arm, replacing the 13 variables in the dynamic graph.

As the Real JACO observations correspond to the generalized coordinates of the simulated JACO Mujoco model, we use the simulated JACO to render the Real JACO trajectories throughout the paper.

\section{Implementation of the models}
\label{sm:sec:implementation}

\subsection{Framework}
\label{sm:sec:framework}

Algorithms were implemented using TensorFlow and Sonnet. We used custom implementations of the graph networks (GNs) as described in the main text.

\subsection{Graph network architectures}
\label{sm:sec:architectures}
Standard sizes and output sizes for the GNs used are:

\begin{itemize}

\item Edge MLP: 2 or 3 hidden layers. 256 to 512 hidden cells per layer.
\item Node and Global MLP: 2 hidden layers. 128 to 256 hidden cells per layer.
\item Updated edge, node and global size: 128
\item (Recurrent models) Node, global and edge size for state graph: 20
\item (Parameter inference) Node, global and edge size for abstract static graph: 10

\end{itemize}

All internal MLPs used layer-wise ReLU activation functions, except for output layers.

\subsection{Data normalization}
\label{sm:sec:normalization}

The two-layer GN core is wrapped by input and output normalization blocks. The input normalization performs linear transformations to produce a zero-mean, unit-variance distributions for each of the global, node and edge features. It is worth noting that for node/edge features, the same transformation is applied to all nodes/edges in the graph, without having specific normalizer parameters for different bodies/edges in the graph. This allows to reuse the same normalizer parameters regardless of the number and type of nodes/edges in the graph. This input normalization is also applied to the observed dynamic graph in the parameter inference network.  

Similarly, inverse normalization is applied to the output nodes of the forward model, to guarantee that the network only needs to output nodes with zero-mean and unit-variance. 

No normalization is applied to the inferred static graph (from the system identification model), in the output of the parameter inference network, nor the input forward prediction network, as in this case the static graph is already represented in a latent feature space.

\subsection{System invariance}
\label{sm:sec:training_invariance}

When training individual models for systems with translation invariance (Swimmer, Cheetah and Walker2d), we always re-centered the system around 0 before the prediction, and moved it back to its initial location after the prediction. This procedure was not applied when multiple systems were trained together.

\subsection{Prediction of dynamic state change}
\label{sm:sec:delta_prediction}
Instead of using the one-step model to predict the absolute dynamic state, we used it to predict the change in dynamic state, which was then used to update the input dynamic state. For the position, linear velocity, and angular velocity, we updated the input by simply adding their corresponding predicted changes. For orientation, where the output represents the rotation quaternion between the input orientation and the next orientation (forced to have unit norm), we computed the update using the Hamilton product.

\subsection{Forward prediction algorithms}

\subsubsection{One-step prediction}
\label{sm:alg:one_step_prediction}

Our forward model takes the system parameters, the system state and a set of actions, to produce the next system state as explained in SM~Algorithm~\ref{sm:alg:forward_algorithm}.

\begin{algorithm}[tb]
   \caption{Forward prediction algorithm.}
   \label{sm:alg:forward_algorithm}
\begin{algorithmic}
   \STATE {\bfseries Input:}  trained GNs $\textrm{GN}_1$, $\textrm{GN}_2$ and normalizers $\textrm{Norm}_{in}$, $\textrm{Norm}_{out}$.
   \STATE {\bfseries Input:} dynamic state $\textbf{x}^{t_0}$ and actions applied $\textbf{x}^{t_0}$ to a system at the current timestep.
   \STATE {\bfseries Input:} system parameters  $\textbf{p}$
   \STATE Build static graph $G_s$ using $\textbf{p}$
   \STATE Build input dynamic nodes $N^{t_0}_d$ using $\textbf{x}^{t_0}$
   \STATE Build input dynamic edges $E^{t_0}_d$ using $\textbf{a}^{t_0}$
   \STATE Build input dynamic graph $G_d$ using $N^{t_0}_d$ and $E^{t_0}_d$
   \STATE Build input graph $G_i = \textrm{concat}(G_s, G_d)$
   \STATE Obtain normalized input graph $G^n_i = \textrm{Norm}_{in}(G_i)$
   \STATE Obtain graph after the first GN:
   $G^{'}= \textrm{GN}_1(G^n_i)$ 
   \STATE Obtain normalized predicted delta dynamic graph:
   $G^{*}= \textrm{GN}_2(\textrm{concat}(G^n_i, G^{'}))$ 
   \STATE Obtain normalized predicted delta dynamic nodes:
   ${\Delta}N^{n}_d= G^{*}\textrm{.nodes}$  
   \STATE Obtain predicted delta dynamic nodes:
   ${\Delta}N_d = \textrm{Norm}^{-1}_{out}({\Delta}N^{n}_d)$
   \STATE Obtain next dynamic nodes $N^{t_0+1}_d$ by updating $N^{t_0}_d$ with ${\Delta}N_d$
   \STATE Extract next dynamic state $\textbf{x}^{t_0+1}$ from $N^{t_0+1}_d$
   \STATE {\bfseries Output:} next system state $\textbf{x}^{t_0+1}$
\end{algorithmic}
\end{algorithm}

\subsubsection{One-step prediction with System ID}
\label{sm:alg:one_step_prediction_sysid}

\begin{algorithm}[tb]
   \caption{Forward prediction with System ID.}
   \label{sm:alg:sysid_forward}
\begin{algorithmic}
   \STATE {\bfseries Input:} trained parameter inference recurrent GN $\textrm{GN}_p$.
   \STATE {\bfseries Input:} trained GNs and normalizers from Algorithm \ref{sm:alg:forward_algorithm}.
   \STATE {\bfseries Input:} dynamic state $\textbf{x}^{t_0}$ and actions applied $\textbf{x}^{t_0}$ to a parametrized system at the current timestep.
   \STATE {\bfseries Input:} a 20-step sequence of observed dynamic states $x^\textrm{seq}$ and actions $x^\textrm{seq}$ for same instance of the system.
   
   \STATE Build dynamic graph sequence $G^\textrm{seq}_d$ using
   $x^\textrm{seq}_i$ and $a^\textrm{seq}_i$
   \STATE Obtain empty graph hidden state $\textrm{G}_\textrm{h}$.
   \FOR{each graph $G^t_d$ in $G^\textrm{seq}_d$}
   \STATE $\textrm{G}_{o},  \textrm{G}_\textrm{h} = \textrm{GN}_p(\textrm{Norm}_{in}(G^t_d), \textrm{G}_\textrm{h})$,
   \ENDFOR
   \STATE Assign $\textrm{G}_{ID} =\textrm{G}_{o}$
   \STATE Use $\textrm{G}_{ID}$ instead of $G_s$ in Algorithm \ref{sm:alg:forward_algorithm} to obtain $\textbf{x}^{t_0+1}$ from $\textbf{x}^{t_0}$ and $\textbf{x}^{t_0}$
   \STATE {\bfseries Output:} next system state $\textbf{x}^{t_0+1}$
\end{algorithmic}
\end{algorithm}

\begin{algorithm}[t]
   \caption{One step of the training algorithm}
   \label{sm:alg:training_one_step}
\begin{algorithmic}
   \STATE {\bfseries Before training:}  initialize weights of GNs $\textrm{GN}_1$, $\textrm{GN}_2$ and accumulators of normalizers $\textrm{Norm}_{in}$, $\textrm{Norm}_{out}$.
   \STATE {\bfseries Input:} batch of dynamic states of the system $\{\textbf{x}^{t_0}\}$ and actions applied $\{\textbf{a}^{t_0}\}$ at the current timestep
   \STATE {\bfseries Input:} batch of dynamic states of the system at the next timestep $\{\textbf{x}^{t_0+1}\}$ 
   \STATE {\bfseries Input:} batch of system parameters  $\{\textbf{p}_i\}$
   \FOR{each example in batch}
   \STATE Build static graph $G_s$ using $\textbf{p}_i$
   \STATE Build input dynamic nodes $N^{t_0}_d$ using $\textbf{x}^{t_0}$
   \STATE Build input dynamic edges $E^{t_0}_d$ using $\textbf{a}^{t_0}$
   \STATE Build output dynamic nodes $N^{t_0+1}_d$ using $\textbf{x}^{t_0+1}$
   \STATE Add noise to input dynamic nodes $N^{t_0}_d$
   \STATE Build input dynamic graph $G_d$ using $N^{t_0}_d$ and $E^{t_0}_d$
   \STATE Build input graph $G_i = \textrm{concat}(G_s, G_d)$
   
   \STATE Obtain target delta dynamic nodes $\Delta N{'}_{d}$ from $N^{t_0+1}_d$ and $N^{t_0}_d$
   \STATE Update $\textrm{Norm}_{in}$ using $G_i$
   \STATE Update $\textrm{Norm}_{out}$ using ${\Delta}N_{d}$
   
   \STATE Obtain normalized input graph $G^n_i = \textrm{Norm}_{in}(G_i)$
   \STATE Obtain normalized target nodes:
   ${\Delta}N^{n'}_d= \textrm{Norm}_{out}({\Delta}N{'}_{d})$
   \STATE Obtain normalized predicted delta dynamic nodes:
   ${\Delta}N^{n}_d= \textrm{GN}_2(\textrm{concat}(G^n_i, \textrm{GN}_1(G^n_i)))\textrm{.nodes}$  
   \STATE Calculate dynamics prediction loss between ${\Delta}N^{n}_d$ and ${\Delta}N^{n'}_d$.
   \ENDFOR
   \STATE Update weights of $\textrm{GN}_1$, $\textrm{GN}_2$ using Adam optimizer on the total loss with gradient clipping.
\end{algorithmic}
\end{algorithm}

For the System ID foward predictions the model takes a system state and a set of actions for a specific instance of a parametrized system, together with a sequence of observed system states and actions for a for the same system instance. The observed sequence is used to identify the system and then produce the next system state as described in Algorithm \ref{sm:alg:sysid_forward}.

In the case of rollout predictions, the System ID is only performed once, on the provided observed sequence, using the same graph for all of the one-step predictions required to generate the trajectory.

\subsection{Training algorithms}
\label{sm:sec:training}

\subsubsection{One-step}
\label{sm:sec:training_one_step}

We trained the one-step forward model in a supervised manner using algorithm \ref{sm:alg:training_one_step}. Part of the training required finding mean and variance parameters for the input and output normalization, which we did online by accumulating information (count, sum and squared sum) about the distributions of the input edge/node/global features, and the distributions of the change in the dynamic states of the nodes, and using that information to estimate the mean and standard deviation of each of the features.

Due to the fact that our representation of the instantaneous state of the bodies is compatible with configurations where the joint constraints are not satisfied, we need to train our model to always produced outputs within the manifold of configurations allowed by the joints. This was achieved by adding random normal noise (magnitude set as a hyper-parameter) to the nodes of the input dynamic graph during training. As a result, the model not only learns to make dynamic predictions, but to put back together systems that are slightly dislocated, which is key to achieve small rollout errors.

\subsubsection{Abstract parameter inference}
\label{sm:sec:training_inference}

\begin{algorithm}[t]
   \caption{End-to-end training algorithm for System ID.}
   \label{sm:alg:sysid_training}
\begin{algorithmic}
   \STATE {\bfseries Before training:} initialize weights of parameter inference recurrent GN $\textrm{GN}_p$, as well as weights from Algorithm \ref{sm:alg:training_one_step}.
   \STATE {\bfseries Input:} a batch of 100-step sequences with dynamic states $\{x^\textrm{seq}_i\}$ and actions $\{x^\textrm{seq}_i\}$
   \FOR{each sequence in batch}
   \STATE Pick a random 20-step subsequence $x^\textrm{subseq}_i$ and $a^\textrm{subseq}_i$.
   \STATE Build dynamic graph sequence $G^\textrm{subseq}_d$ using
   $x^\textrm{subseq}_i$ and $a^\textrm{subseq}_i$
   \STATE Obtain empty graph hidden state $\textrm{G}_\textrm{h}$.
   \FOR{each graph $G^t_d$ in $G^\textrm{subseq}_d$}
   \STATE $\textrm{G}_{o},  \textrm{G}_\textrm{h} = \textrm{GN}_p(\textrm{Norm}_{in}(G^t_d), \textrm{G}_\textrm{h})$,
   \ENDFOR
   \STATE Assign $\textrm{G}_{ID} =\textrm{G}_{o}$
   \STATE Pick a different random timestep $t_0$ from  $\{x^\mathrm{seq}_i\}$, $\{x^\textrm{seq}_i\}$ 
   \STATE Apply Algorithm \ref{sm:alg:training_one_step} to timestep $t_0$ using final $\textrm{G}_{ID}$ instead $G_s$ to obtain the dynamics prediction loss.
   \ENDFOR
   \STATE Update weights of $\textrm{GN}_p$, $\textrm{GN}_1$, $\textrm{GN}_2$ using Adam optimizer on the total loss with gradient clipping.
\end{algorithmic}
\end{algorithm}

The training of the parameter inference recurrent GN is performed as described in Algorithm \ref{sm:alg:sysid_training}. The recurrent GN and the dynamics GN are trained together end-to-end by sampling a random 20-step sequence for the former, and a random supervised example for the latter from 100-step graph sequences, with a single loss based on the prediction error for the supervised example. This separation between the sequence at the supervised sample, encourages the recurrent GN to truly extract abstract static properties that are independent from the specific 20-step trajectory, but useful for making dynamics predictions under any condition.

\subsubsection{Recurrent one-step predictions}
\label{sm:sec:training_recurrent}

The one-step prediction recurrent model, used for the Real JACO predictions, is trained from 21-step sequences using the \emph{teacher forcing} method. The first 20 graphs in the sequence are used as input graphs, while the last 20 graphs in the sequence are used as target graphs. During training, the recurrent model is used to sequentially process the input graphs, producing at each step a predicted dynamic graph, which is stored, and a graph state, which is fed together with the next input graph in the next iteration. After processing the entire sequence, the sequence of predicted dynamic graphs and the target graphs are used together to calculate the loss.

\subsubsection{Loss}
\label{sm:sec:training_loss}
We use a standard L2-norm between the normalized expected and predicted delta nodes, for the position, linear velocity, and angular velocity features. We do this for the normalized features to guarantee a balanced relative weighting between the different features. In the case of the orientation, we cannot directly calculate the L2-norm between the predicted rotation quaternion $\textbf{q}_p$ to the expected rotation quaternion $\textbf{q}_e$, as a quaternion $\textbf{q}$ and $-\textbf{q}$ represent the same orientation. Instead, we minimize the angle distance between $\textbf{q}_p$ and $\textbf{q}_e$ by minimizing the loss $1-\cos^2{(\textbf{q}_e\cdotp\textbf{q}_p)}$ after.

\subsection{Training details}
\label{sm:sec:training_details}

Models were trained with a batch size of 200 graphs/graph sequences, using an Adam optimizer on a single GPU. Starting learning rates were tuned at $1^{-4}$. We used two different exponential decay with factor of 0.975 updated every 50000 (fast training) or 200000  (slow training) steps.

We trained our models using early stopping or asymptotic convergence based the rollout error on 20-step sequences from the validation set. Simple environments (such as individual fixed environments) would typically train using the fast training configuration for a period between less than a day to a few days, depending on the size of the environment and the size of the network. Using slow training in these cases only yields a marginal improvement. On the other hand, more complex models such as those learning multiple environments and multiple parametrized environments benefited from the slow training to achieve optimal behavior for periods of between 1-3 weeks.

\section{MLP baseline architectures}
\label{sm:sec:MLP baseline}

For the MLP baselines, we used 5 different models (ReLU activation) spanning a large capacity range:
\begin{itemize}
    \item 3 hidden layers, 128 hidden cells per layer
    \item 3 hidden layers, 512 hidden cells per layer
    \item 9 hidden layers, 128 hidden cells per layer
    \item 9 hidden layers, 512 hidden cells per layer
    \item 5 hidden layers, 256 hidden cells per layer
\end{itemize}

The corresponding MLP replaces the 2-layer GN core, with additional layers to flatten the input graphs into feature batches, and to reconstruct the graphs at the output. Both normalization and graph update layers are still applied at graph level, in the same way that for the GN-based model.

Each of the models was trained four times using initial learning rates of $1^{-3}$ and $1^{-4}$ and learning rate decays every 50000 and 200000 steps. The model performing best on validation rollouts for each environment, out of the 20 hyperparameter combinations was chosen as the MLP baseline.

\section{Control}
\label{sm:sec:control}

\subsection{Model-based planning algorithms}
\label{sm:sec:model_based_planning}

\subsubsection{MPC planner with learned models}
\label{sm:sec:planning_mpc}

\begin{algorithm}[tb]
   \caption{MPC algorithm}
   \label{sm:alg:MPC}
\begin{algorithmic}
   \STATE {\bfseries Input:} initial system state $\textbf{x}^{0}$,
   \STATE {\bfseries Input:} randomly initialized sequence of actions $\{\textbf{a}^{t}\}$.
   \STATE {\bfseries Input:} pretrained dynamics model $M$ such
   
   $\textbf{x}^{t_0+1} = M(\textbf{x}^{t_0}, \textbf{a}^{t_0})$
   \STATE {\bfseries Input:}  Trajectory cost function $L$ such
   
   $c = C(\{\textbf{x}^{t}\}, \{\textbf{a}^{t}\})$
   \FOR{a number of iterations}
   \STATE $\textbf{x}^{0}_r=\textbf{x}^{0}$
   \FOR{t in range(0, horizon)}
     \STATE $\textbf{x}^{t+1}_r=M(\textbf{x}^{t}_r, \textbf{a}^{t})$
   \ENDFOR
   \STATE Calculate trajectory cost $c = C(\{\textbf{x}_r^{t}\}, \{\textbf{a}^{t}\})$
   \STATE Calculate gradients $\{\textbf{g}_a^{t}\} = \frac{\partial c}{\partial \{\textbf{a}^{t}\}}$
   \STATE Apply gradient based update to $\{\textbf{a}^{t}\}$
   \ENDFOR
   \STATE {\bfseries Output:} optimized action sequence $\{\textbf{a}^{t}\}$ 
\end{algorithmic}
\end{algorithm}

We implemented MPC using our learned models as explained in SM~Algorithm~\ref{sm:alg:MPC}. We applied the algorithm in a receding horizon manner by iteratively planning for a fixed horizon (see SM~Table~\ref{sm:sec:planning_configuration}), applying the first action of the sequence, and increasing the horizon by one step, reusing the shifted optimal trajectory computed in the previous iteration. We typically performed between 3 and 51 optimization iterations $N$ from each initial state, with additional $N\cdot \textrm{horizon}$ iterations at the very first initial state, to \emph{warm-up} the fully-random initial action sequence.

\subsubsection{Baseline Mujoco-based planner}
\label{sm:sec:planning_mujoco_baseline}

As a baseline planning approach we used the iterative Linear-Quadratic-Gaussian (iLQG) trajectory optimization approach proposed in \cite{tassa2014control}.  This method alternates between forward passes (rollouts) which integrate the dynamics forward for a current control sequence and backwards passes which consists of perturbations to the control sequence to improve upon the recursively computed objective function.  
Note that in the backwards pass, each local perturbation can be formulated as an optimization problem, and linear inequality constraints ensure that the resulting control trajectory does not require controls outside of the range that can be feasibly generated by the corresponding degrees of freedom in the MuJoCo model.
The overall objective optimized corresponds to the total cost over $J$ a finite horizon:
\begin{equation}
    J(x_0, U) = \sum_{t=0}^{T-1} \ell (x_t, u_t) + \ell (x_T)
\end{equation}
where $x_0$ is the initial state, $u_t$ is the control signal (i.e. action) taken at timestep $t$, $U$ is the trajectory of controls, $\ell(\cdot)$ is the cost function.  We assume the dynamics are deterministic transitions $x_{t+1} = f(x_t, u_t)$. 

While this iLQG planner does not work optimally when the dynamics involve complex contacts, for relatively smooth dynamics as found in the swimmer, differential dynamic programming (DDP) style approaches works well \cite{tassa2008receding}.  Relevant cost functions are presented in SM~Section~\ref{sm:sec:planning_configuration}. 

\subsection{Planning configuration}
\label{sm:sec:planning_configuration}

\vskip 0.15in
\begin{center}
\begin{small}
\renewcommand{\arraystretch}{1.5}
\begin{tabular}{p{15mm}p{30mm}p{15mm}p{90mm}}
\toprule
 Name & Task & Planning \newline horizon & Reward to maximize (summed for all timesteps) \\
\midrule
Pendulum  & Balance & 50, 100 & Negative angle between the quaternion of the pendulum and the target quaternion corresponding to the balanced position. (0 when balanced at the top, $<0$ otherwise).
\\
Cartpole  & Balance & 50, 100 & Same as Pendulum-Balance calculated for the pole.\\
Acrobot   & Swing up & 100 & Same as Pendulum-Balance summed for both acrobot links.\\
Swimmer   & Mover towards target & 100 & Projection of the displacement vector of the Swimmer head from the previous timestep on the target direction, The target direction is calculated as the vector joining the head of the swimmer at the first planning timestep with the target location. The reward is shaped (0.01 contribution) with the negative squared projection on the perpendicular target direction.\\
Cheetah   & Move forward & 20 & Horizontal component of the absolute velocity of the torso.\\
   & Vertical position & 20 & Vertical component of the absolute position of the torso.\\
   & Squared vertical speed &20 & Squared vertical component of the absolute velocity of the torso.\\
   & Squared angular speed & 20& Squared angular velocity of the torso.\\
Walker2d   & Move forward &20 & Horizontal component of the absolute velocity of the torso.\\
   & Vertical position &20 & Vertical component of the absolute position of the torso.\\
   & Inverse verticality &20& Same as Pendulum-Balance summed for torso, thighs and legs.\\
   & Feet to head height &20& Summed squared vertical distance between the position of each of the feet and the height of Walker2d.\\
Jaco      & Imitate Palm Pose &20& Negative dynamic-state loss (as described in Section \ref{sm:sec:training_loss}) between the position-and-orientation of the body representing the JACO palm and the target position-and-orientation .\\
  & Imitate Full Pose &20& Same as Jaco-Imitate Palm Pose but summed across all the bodies forming JACO (see SM~Section~\ref{sm:sec:training_loss}).\\

\bottomrule
\end{tabular}
\end{small}
\end{center}
\vskip -0.1in
\renewcommand{\arraystretch}{1.}

\subsection{Reinforcement learning agents}
\label{sm:sec:reinforcement_learning}
Our RL experiments use three base algorithms for continuous control: DDPG \citep{Lillicrap16}, SVG(0) and SVG(N)~\citep{heess2015learning}. All of these algorithms find a policy $\pi$ that selects an action $a$ in a given state $x$ by maximizing the expected discounted reward,
\begin{equation}
Q(\mathbf{x}, \mathbf{a}) = \mathbb{E} \Big[ \sum_{t=0}^\infty \gamma^t r(\mathbf{x}, \mathbf{a}) \Big],
\end{equation}
where $r(x,a)$ is the per-step reward and $\gamma$ denotes the discount factor.
Learning in all algorithms we consider occurs off-policy. That is, we continuously generate experience via the current best policy $\pi$, storing all experience (sequences of states, actions and rewards) it into a replay buffer $\mathcal{B}$, and minimize a loss defined on samples from $\mathcal{B}$ via stochastic gradient descent.

\subsubsection{DDPG}
\label{sm:sec:ddpg}
The DDPG algorithm~\citep{Lillicrap16} learns a deterministic control policy $\pi = \mu_\theta(s)$ with parameters $\theta$ and a corresponding action-value function $Q_\phi^\mu(s, a)$, with parameters $\phi$. Both of these mapping are parameterized via neural networks in our experiments.

Learning proceeds via gradient descent on two objectives. The objective for learning the $Q$ function is to minimize the one-step Bellman error using samples from a replay buffer, that is we seek to find $\arg \min_\phi L(\phi)$ by following the gradient,
\begin{equation}
\begin{aligned}
\nabla_\phi L(\phi) &= \mathbb{E}_{(\mathbf{x}_t, \mathbf{a}_t, \mathbf{x}_{t+1}, r_t) \in \mathcal{B}} \Big[ \nabla_\phi \big( Q^\mu_\phi(\mathbf{x}_t, \mathbf{a}_t) - y \big)^2\Big], \\
\text{with } \ y &= r_t + \gamma Q^\mu_{\phi'}(\mathbf{x}_{t+1}, \mu_{\theta'}(\mathbf{x}_{t+1}))
\end{aligned}
\label{eq:q_learning}
\end{equation}
where $\phi'$ and $\theta'$ denote the parameters of target Q-value and policy networks, that are periodically copied from the current parameters, this is common practice in RL to stabilize training (we update the target networks every $1000$ gradient steps).
The objective for learning the policy is performed by searching for an action that obtains maximum value, as judged by the learned Q-function. That is we find $\arg \min_\theta L(\theta)$ by following the deterministic policy gradient \cite{Lillicrap16},
\begin{equation}
\nabla_\theta L^{\text{DPG}}(\theta) = \mathbb{E}_{\mathbf{x}_t \in \mathcal{B}} \Big[ - \nabla_\theta Q^\mu_\theta(\mathbf{x}_t, \mu_\theta(\mathbf{x}_t))\Big].
\label{eq:ddpg_grad}
\end{equation}

\subsubsection{SVG}
\label{sm:sec:svg}
For our experiments with the familiy of Stochastic Value Gradient (SVG) \citep{heess2015learning} algorithms we considered two-variants a model-free baseline SVG(0) that optimizes a stochastic policy based on a learned Q-function as well as a model-based version SVG(N) (using our Graph Net model) that unrolls the system dynamics for N-steps.

\paragraph{SVG(0)} In the model-free variant learning proceeds similarly to the DDPG algorithm. We learn both, a parametric Q-value estimator as well as a (now stochastic) policy $\pi_\theta(\mathbf{a} | \mathbf{x})$ from which actions can be sampled. 
In our implementation learning of the Q-function is performed by following the gradient from Equation \eqref{eq:q_learning}, with $\mu(x)$ replaced by samples $\mathbf{a} \sim \pi_\theta(\mathbf{a} | \mathbf{x})$. 

For the policy learning step we can learn via a stochastic analogue of the deterministic policy gradient from Equation \eqref{eq:ddpg_grad}, the so called stochastic value gradient, which reads
\begin{equation}
\nabla_\theta L^{\text{SVG}}(\theta) = - \nabla_\theta \mathbb{E}_{\substack{\mathbf{x}_t \in \mathcal{B} \\ \mathbf{a} \sim \pi_\theta(\mathbf{a} | \mathbf{x}_t)}} \Big[ Q^\pi_\theta(\mathbf{x}_t, \cdot)\Big].
\label{eq:svg0}
\end{equation}
For a Gaussian policy (as used in this paper) the gradient of this expectation can be calculated via the reparameterization trick \citep{Kingma2013auto,rezende14}.

\paragraph{SVG(N)}
For the model based version we used a variation of SVG(N) that employs an action-value function -- instead of the value function estimator used in the original paper. This allowed us to directly compare the performance of a SVG(0) agent, which is model free, with SVG(1) which calculates policy gradients using a one-step model based horizon. 

In particular, similar to Equation \eqref{eq:svg0}, we obtain the model based policy gradient as 
\begin{equation}
\nabla_\theta L^{\text{SVG(N)}}(\theta) = - \nabla_\theta \mathbb{E}_{
\substack{\mathbf{x}_t \in \mathcal{B} \\ 
 \mathbf{a}_t \sim \pi_\theta(\mathbf{a} | \mathbf{x}_t)\\ 
 \mathbf{a}_{t+1} \sim \pi_\theta(\mathbf{a} | \mathbf{x}_{t+1})}} \Big[ r_{t}(\mathbf{x}_t, \mathbf{a}_t) + \gamma Q^\pi_\theta(\mathbf{x}_{t+1}, \mathbf{a}_t) \mid \mathbf{x}_{t+1} = g(\mathbf{x}_t, \mathbf{a}_t) \Big],
\end{equation}
where $g$ denotes the dynamics, as predicted by the GN and the gradient can, again, be computed via reparameterization (we refer to \citet{heess2015learning} for a detailed discussion).

\begin{figure}[ht]
\vskip 0.2in
\begin{center}
\includegraphics[width=0.6\columnwidth]{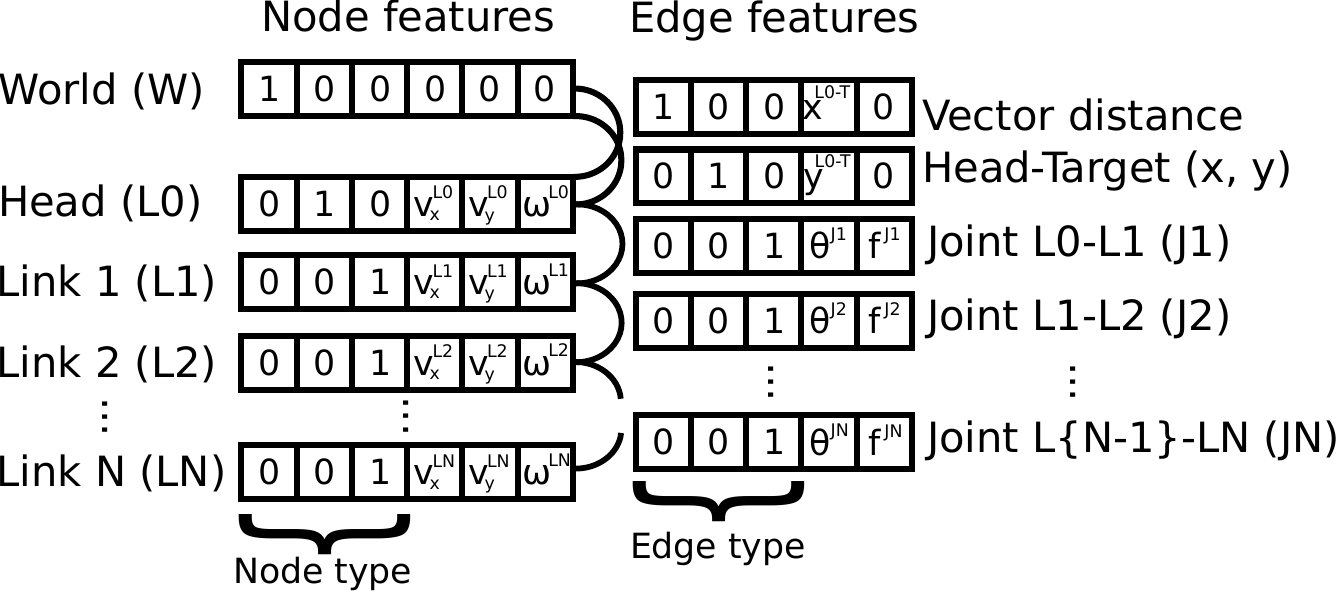}
\caption{Arrangement as a graph of the default 25-feature observation and 5 actions provided in the Swimmer 6 task from the DeepMind Control Suite \cite{tassa2018deepmind}. The observation consists of: \emph{(to\_target)} the distance between the head and the target projected in the axis of the head ($x^{\text{L0-T}}$, $y^{\text{L0-T}}$), \emph{(joints)} the angle of each joint JN between adjacent swimmer links L{N-1} and L{N} ($\theta^{\text{JN}}$) and \emph{(body\_velocities)} the linear and angular velocity of each link LN projected in its own axis ($v_x^{\text{LN}}$, $v_y^{\text{LN}}$, $\omega^{\text{LN}}$). The actions consists of the force applied to each of the joints ($f^\text{JN}$) connecting the links. Because our graphs are directed, all of the edges were duplicated, with an additional {-1, 1} feature indicating the direction.}
\label{sm:fig:swimmer_graph_conversion}
\end{center}
\vskip -0.2in
\end{figure}

We experimented with SVG(1) on the swimmer domain with six links (Swimmer 6). Since in this case, the goal for the GN is to predict environment observations (as opposed to the full state for each body), we constructed a graph from the observations and actions obtained from the environment.
SM Figure \ref{sm:fig:swimmer_graph_conversion} describes the observations and actions and shows how they were transformed into a graph.

\section{Mujoco variables included in the graph conversion}
\label{sm:sec:mujoco_variables}

\subsection{Dynamic graph}

We retrieved the the absolute position, orientation, linear and angular velocities for each body:
\begin{itemize}
\item \textbf{Global: None}
\item \textbf{Nodes:} (for each body)\\
Absolute body position (3 vars): mjData.xpos \\
Absolute body quaternion orientation position (4 vars): mjData.xquat \\
Absolute linear and angular velocity (6 vars): mj\_objectVelocity
(mjOBJ\_XBODY, flg\_local=False)
\item \textbf{Edges:} (for each joint) Magnitude of action at joint: mjData.ctrl (0, if not applicable).
\end{itemize}

\subsection{Static graph}
We performed an exhaustive selection of global, body, and joint static properties from mjModel:
\begin{itemize}
\item \textbf{Global:} mjModel.opt.\{timestep, gravity, wind, magnetic, density, viscosity, impratio, o\_margin, o\_solref, o\_solimp, collision\_type (one-hot), enableflags (bit array), disableflags (bit array)\}.
\item \textbf{Nodes:} (for each body) mjModel.body\_\{mass, pos, quat, inertia, ipos, iquat\}.
\item \textbf{Edges:} (for each joint) \\Direction of edge (1: parent-to-child, -1: child-to-parent). \\Motorized flag (1: if motorized, 0 otherwise). \\Joint properties: mjModel.jnt\_\{type (one-hot), axis, pos, solimp, solref, stiffness, limited, range, margin\}. \\Actuator properties: mjModel.opt.actuator\_\{biastype (one-hot), biasprm, cranklength, ctrllimited, ctrlrange, dyntype (one-hot), dynprm, forcelimited, forcerange, gaintype (one-hot), gainprm, gear, invweight0, length0, lengthrange\}.
\end{itemize}

Most of these properties are constant for all environments use, however, they are kept for completeness. While we do not include geom properties such as size, density or shape, this information should be partially encoded in the inertia tensor together with the mass.

\section{Supplementary figures}
\label{sm:sec:supplementary_figures}

\begin{figure}[ht]
\vskip 0.2in
\begin{center}
\includegraphics[width=0.45\columnwidth]{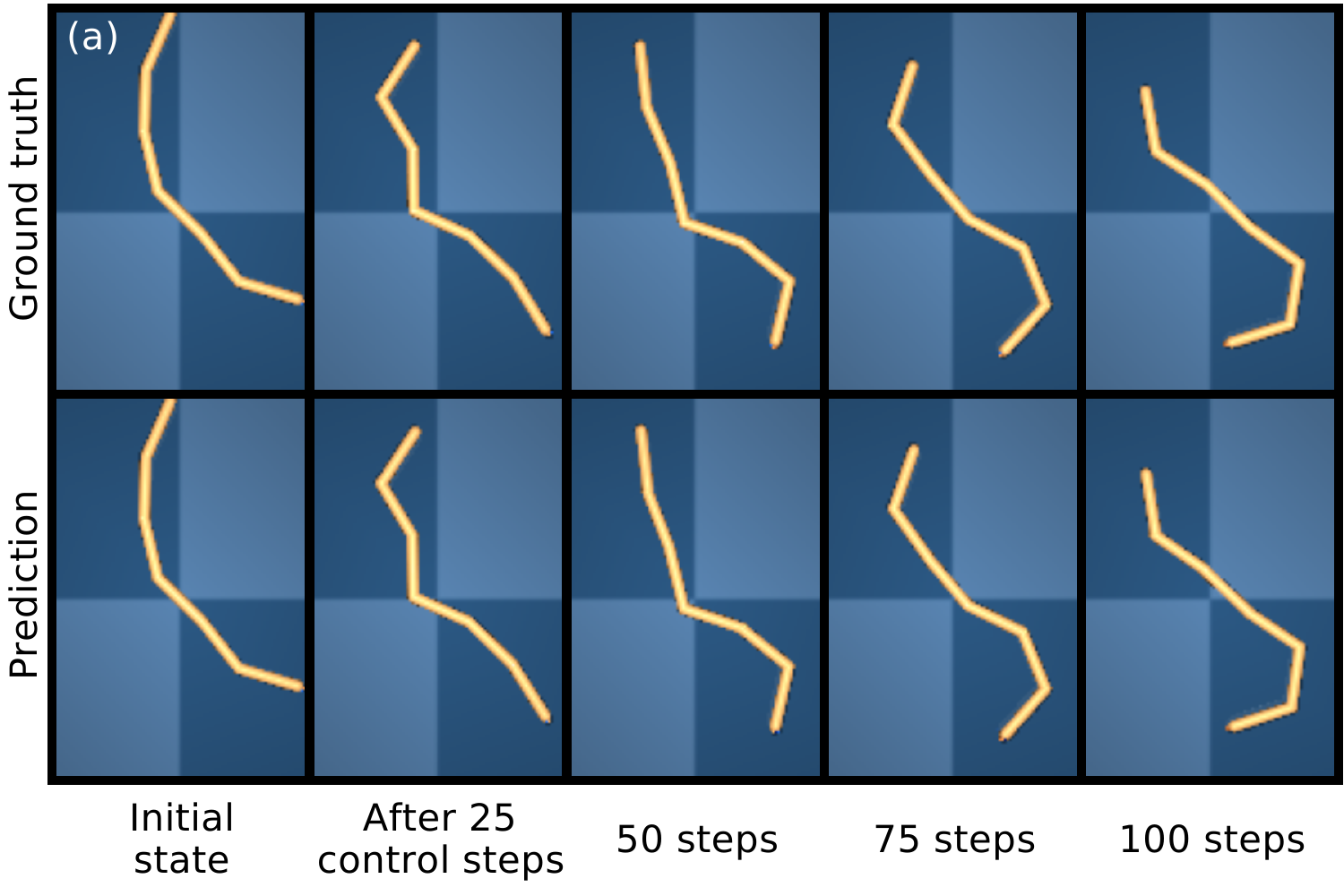}
\hspace{0.2cm}
\includegraphics[width=0.46\columnwidth]{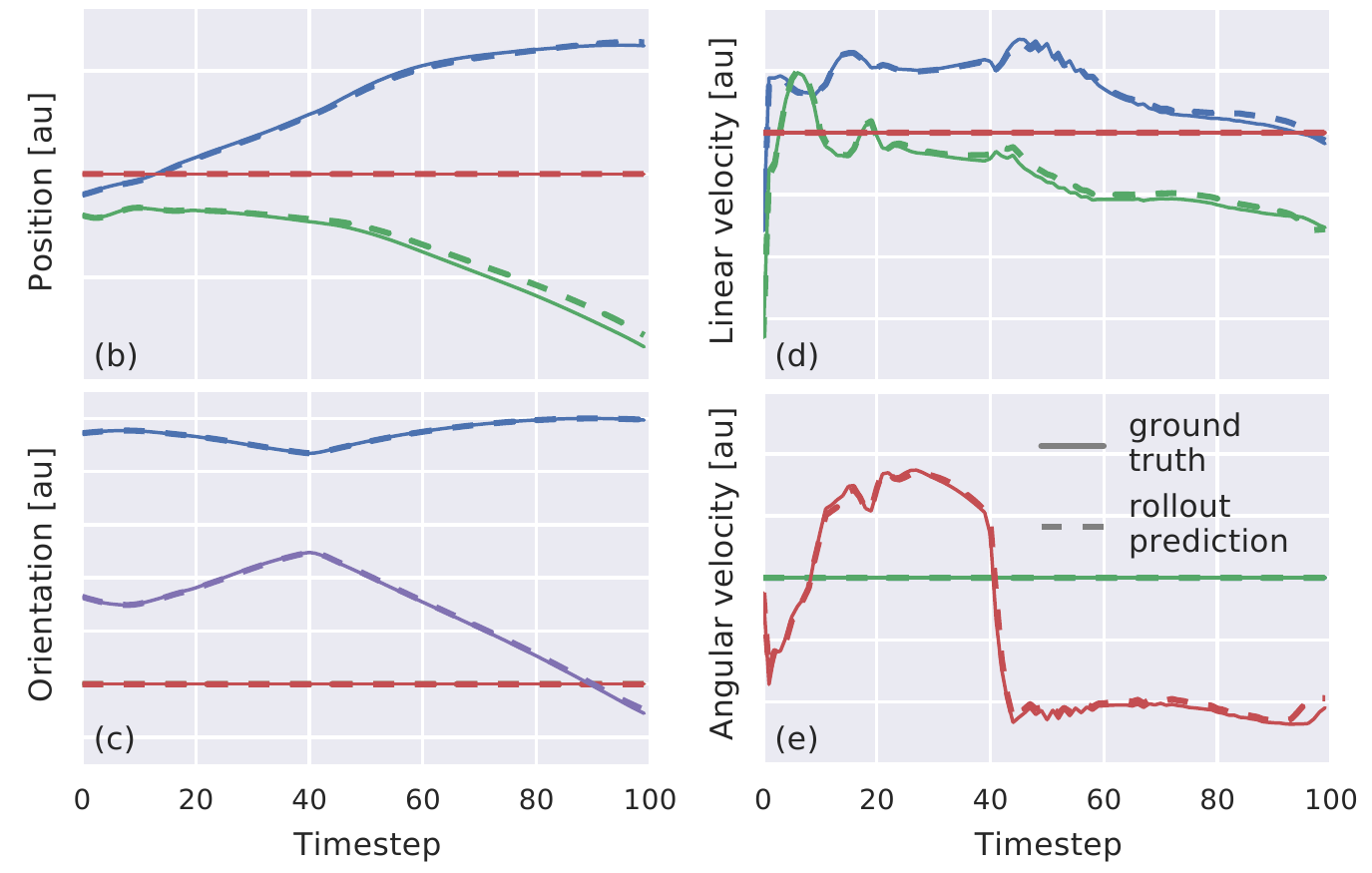}
\caption{Model trained on Swimmer6 trajectories under random control evaluated on a trajectory generated by a DDPG agent. Trajectories are also available in video [\href{\rolloutsswimmerdpgdata}{\rolloutsswimmerdpgdataname}]. 
(Left) Key-frames comparing the ground truth and predicted sequence within a 100 step trajectory. (Right) Full state sequence prediction for the third link of the Swimmer, consisting of Cartesian position (3 vars), quaternion orientation (4 vars), Cartesian linear velocity (3 vars) and Cartesian angular velocity (3 vars). The full prediction contains such 13 variables for each of the links, that is 78 variables.
}
\label{sm:fig:swimmer_rollout_dpg}
\end{center}
\vskip -0.2in
\end{figure}

\begin{figure}[ht]
\vskip 0.2in
\begin{center}
\centerline{\includegraphics[width=0.5\columnwidth]{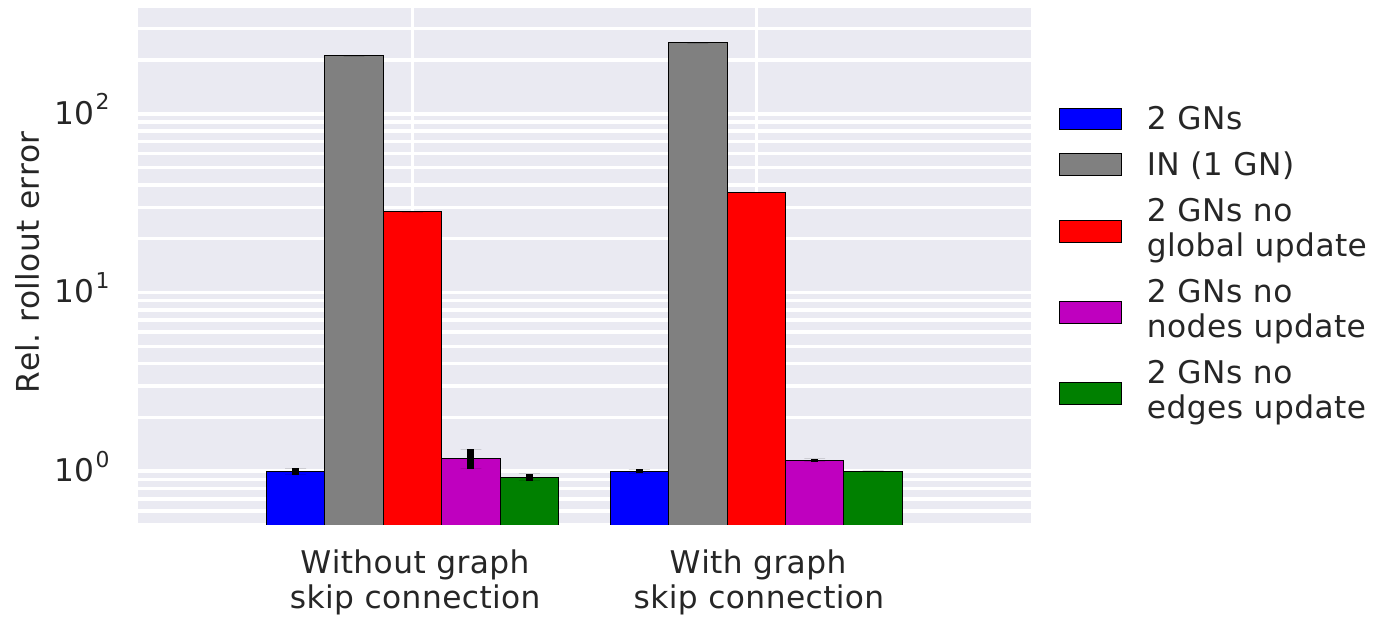}}
\caption{Ablation study of the architecture using the rollout error over 20 step test sequences in Swimmer6 to evaluate relative performance. The performance of the architecture used in this work (a sequence of two GNs, blue) is compared to: an Interaction Network (IN) \cite{battaglia2016interaction}, which is equivalent in this case to a single GN (grey), and a sequence of GNs where the first GN is not allowed to update either the global (red), the nodes (purple) or the edges(green) of the output graph. Results are shown both for a purely sequential connection of the GNs, and for a model with a graph skip connection, where the output graph of the first GN, is concatenated to the input graph, before feeding it into the second GN. The results show that the performance of the double GN is far superior than that of the equivalent IN. They also show that the global update performed by the GN is necessary for the model to perform well. We hypothesize this is due to the long range dependencies within the graph that exist within swimmer, and the ability of the global update to quickly propagate such dependencies across the entire graph. Similar results may have been obtained without global updates by using a deeper stack of GNs to allow information to flow across the entire graph.
Each model was trained from three different seeds. The figure depicts the mean, and the standard deviation of the asymptotic performance of the three seeds.}
\label{sm:fig:ablation_study}
\end{center}
\vskip -0.2in
\end{figure}

\end{document}